\documentclass[%
 reprint,
 amsmath,amssymb,
 aps,
]{revtex4-2}

\usepackage{graphicx}
\usepackage{dcolumn}
\usepackage{bm}

\usepackage{booktabs}
\usepackage{arydshln}
\begin{document}

\preprint{APS/123-QED}

\title{Exploiting Chaotic Dynamics as Deep Neural Networks}


\author{
Shuhong Liu${}^{1*}$,
Nozomi Akashi${}^{2}$, 
Qingyao Huang${}^{1}$, \\
Yasuo Kuniyoshi${}^{1}$}
\author{
Kohei Nakajima${}^{1}$}\altaffiliation{To whom correspondence should be addressed. \\ E-mail: \{s-liu, k-nakajima\}@isi.imi.i.u-tokyo.ac.jp
}

\affiliation{
\vspace{0.5cm}
\normalsize{${}^{1}$Graduate School of Information Science and Technology, The University of Tokyo}\\
\normalsize{7-3-1 Hongo, Bunkyo-ku, 113-8656 Tokyo, Japan}\\
\normalsize{${}^{2}$Graduate School of Informatics, Kyoto University} \\
\normalsize{Yoshida-honmachi, Sakyo-ku, 606-8501 Kyoto, Japan}\\
}

\begin{abstract}
Chaos presents complex dynamics arising from nonlinearity and a sensitivity to initial states. These characteristics suggest a depth of expressivity that underscores their potential for advanced computational applications. However, strategies to effectively exploit chaotic dynamics for information processing have largely remained elusive. In this study, we reveal that the essence of chaos can be found in various state-of-the-art deep neural networks. Drawing inspiration from this revelation, we propose a novel method that directly leverages chaotic dynamics for deep learning architectures. Our approach is systematically evaluated across distinct chaotic systems. In all instances, our framework presents superior results to conventional deep neural networks in terms of accuracy, convergence speed, and efficiency. Furthermore, we found an active role of transient chaos formation in our scheme. Collectively, this study offers a new path for the integration of chaos, which has long been overlooked in information processing, and provides insights into the prospective fusion of chaotic dynamics within the domains of machine learning and neuromorphic computation.
\end{abstract}

\maketitle


\section{Introduction}
Chaos exists ubiquitously in nature. It is characterized as a bounded and deterministic system with a high sensitivity to small initial state perturbations. As time evolves, small perturbations in the system can yield exponentially expanding outcomes. Chaotic behaviors are prevalent across domains, from the intricacies of atmospheric dynamics to fluid mechanics \cite{lorenz:1963:deterministic}. Chaos is also evident in engineered systems, from the simple double-pendulum system to complex neuromorphic devices \cite{yang:2007:chaotic}. Additionally, numerous studies have illuminated the presence of chaotic dynamics in biological neural systems \cite{schiff:1994:controlling,korn:2003:there}, evident in areas from electroencephalogram signal analysis \cite{pritchard:1995:measuring} to the dynamical behaviors observed in resting-state brain activity \cite{deco:2011:emerging}.

The adaptability, diversity, and energy efficiency inherent to biological neural systems offer compelling blueprints for novel computational possibilities. Drawing inspiration from these, recent advancements in neuromorphic computing offer the potential to overcome the speed limit due to the von Neumann bottleneck \cite{backus:1978:can}. Various neuromorphic devices have been proposed, employing distinct modalities such as spintronics, photonics, quantum, or nanomaterials \cite{markovic:2020:physics}. A particular benefit lies in the inherent dynamics of these physical systems, which facilitate proactive state evolution without requiring additional computation overhead. Such natural evolution of physical components over time ensures computation with minimal external intervention, endowing neuromorphic devices with exceptional computation and energy efficiency.

Despite the appeal of neuromorphic systems, integrating them with chaotic dynamics has posed challenges. Reservoir computing (RC) \cite{jaeger:2004:harnessing, maass:2002:real, nakajima:2021:reservoir} has achieved significant success in implementing neuromorphic systems. RC employs inherent dynamics, termed the ``reservoir'', as a recurrent neural network (RNN), enabling efficient hardware implementation and training processes. However, an important condition for achieving reliable performance in RC is the echo-state property (ESP) \cite{jaeger:2004:harnessing}. This property guarantees that the reservoir's internal state eventually attains independence from its initial conditions, ensuring reproducible computations. This demand straightforwardly contradicts the intrinsic property of chaos, which typically exhibits sensitivity to initial conditions. Recent discoveries present the clue of expansion behavior, a crucial characteristic of chaos, in artificial deep neural networks (DNNs) \cite{ishihara:2005:magic,saxe:2013:exact,mototake:2015:dynamics,schoenholz:2016:deep,poole:2016:exponential,lin:2020:chaotic,keup:2021:transient,inoue:2022:transient,engelken:2023:lyapunov}, suggesting a profound interplay between chaos and information processing. 

\begin{figure*}[t!]
	\begin{center}
		\includegraphics[width=11.4cm]{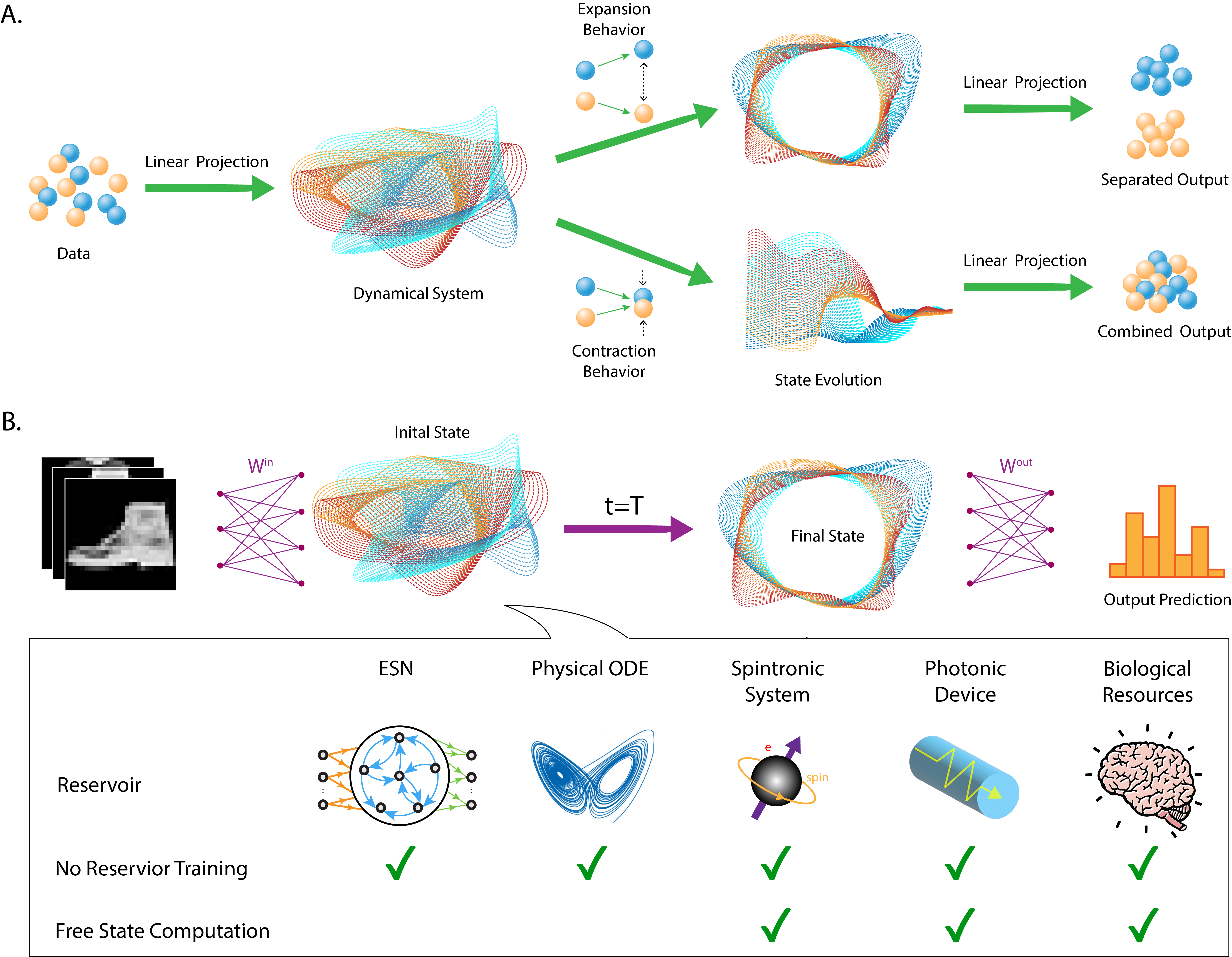}
	\end{center}
	\caption{\textbf{A.} The illustration of the expansion behavior within the dynamical system. \textbf{B.} Conceptual design of the proposed framework that utilizes chaotic dynamics for the purpose of image classification. The terms $W_{\rm in}$ and $W_{\rm out}$ denote the linear transformation of the state space, while $T$ signifies the duration of self-evolution in the employed dynamical system.}
	\label{fig:overview}
\end{figure*}

Prompted by these insights, our study dives deep into the heart of the challenge: Can chaotic dynamics truly be assets in DNNs? Our research systematically examines the proficiency of these dynamics in information processing, challenging conventional notions that view such dynamics as non-contributive. For DNNs, our investigation centers on revealing the existence of expansion properties. Our observations reveal the essence of chaos in state-of-the-art DNNs and illuminate its pivotal role in input separation. With this finding, we then propose a straightforward but effective framework that utilizes chaotic dynamics directly as computational mediums. This strategy requires much fewer trainable weights while exhibiting rapid convergence. The implications of our approach are particularly profound for the future of neuromorphic devices. To underscore its effectiveness, our framework is assessed across diverse chaotic dynamics—from discrete to continuous systems and from spatiotemporal chaos to intricate neuromorphic systems based on spin-torque oscillators. 

\section{Results}

\label{results}
Given the lack of (i) a general assessment of chaotic properties in DNNs (detailed in Appendix \ref{sup:chaos_in_dnns}) and (ii) efficient methods for utilizing chaotic dynamics for information processing, our study unfolds in two parts. Initially, we analyze the behavior of conventional DNNs from a dynamical system's viewpoint, focusing on examining the emergence of the expansion property in various tasks. Subsequently, we propose a novel framework for exploiting chaotic dynamics as the computational medium and conducted experiments to substantiate its effectiveness. Figure~\ref{fig:overview}A depicts the fundamental characteristics of expansion property. In this process, input data from closely situated positions clearly separate through diverging paths as the state evolves. Figure~\ref{fig:overview}B outlines the conceptual design of our proposed framework. This design capitalizes on the intrinsic dynamics of the integrated system, wherein only two layers of linear projection can be trained. This approach, especially when incorporating physical systems, significantly enhances energy efficiency by leveraging natural state progression.

\begin{figure*}[t!]
\centering
    \includegraphics[width=17.8cm]{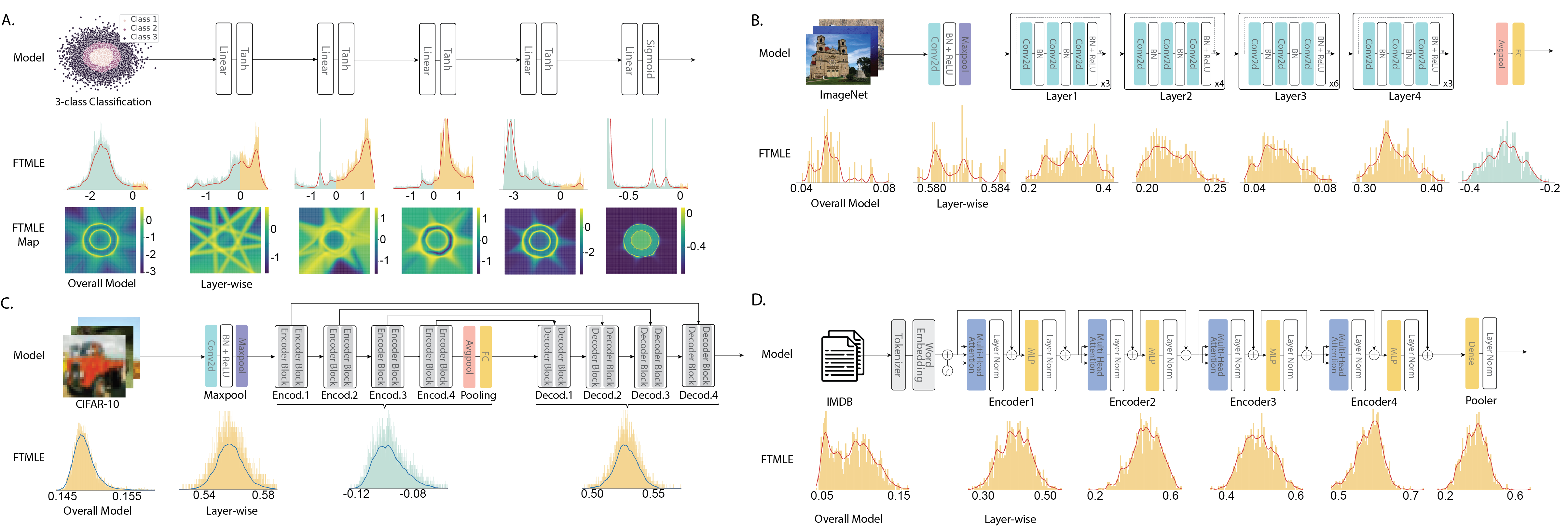}
\caption{FTMLE analyses for various deep neural networks. The FTMLE distribution corresponds with overall or layer-wise FTMLE values with respect to the input samples. Within the diagrams, blue areas denote contraction zones, while orange areas signify expansion zones. \textbf{A.} Five-layer MLP trained for a three-class classification task in a two-dimensional state space. The FTMLE analysis is conducted on the testing set uniformly sampled from the state space. The FTMLE map reflects the corresponding FTMLE values of the sampled test data. \textbf{B.} FTMLE analysis on a pre-trained ResNet50 model using a subset of the ImageNet validation set, where one image is sampled for each of the 1,000 classes. \textbf{C.} Pretrained ResNet-based auto-encoder evaluated on CIFAR10 testing set. \textbf{D.} Pretrained BERT-mini evaluated on IMDb testing set.}
\label{fig:DNN-FTMLE}
\end{figure*}

\subsection{Expansion Property in Deep Neural Networks}
To reveal the presence of expansion property in neural networks, we chose a variety of state-of-the-art architectures, including the multilayer perceptron (MLP), convolutional neural network (CNN), convolutional auto-encoder (AE), and transformer-based language model known as Bidirectional Encoder Representations from Transformers (BERT), and applied them to tasks across distinct domains. We analyze the model from two perspectives---its overall model and layer-wise dynamics---by assessing their transient dynamics using the finite-time maximum Lyapunov exponent (FTMLE), which can be derived from a general form of Lyapunov exponent

\begin{align}
    \lambda[T]({\bf x}_i) &= \max_{\delta {\bf x}_i}{\frac{1}{T}\log{\frac{\parallel \delta {\bf x}_{i+T} \parallel}{\parallel \delta {\bf x}_i \parallel}}} \label{eq:general_LE}
\end{align}

\noindent by using a finite $T$ as

\begin{align}
    \lambda_{T}^{\rm FTMLE}({\bf x}_i) &= \lambda[T]({\bf x}_i) ~, \label{eq:FTMLE}
\end{align}

\noindent where $x_i$ represents the current state at time $i$, $\delta x_i$ indicates the perturbation applied on $x_i$, and $T$ is typically the time of iteration. In the context of evaluating FTMLE in DNNs, $T$ denotes the number of propagated layers. the maximum Lyapunov exponent (MLE), as outlined in Equation \ref{eq:MLE} in the Methods section, characterizes the global dynamics of the system as $T$ tends towards infinity. Conversely, the local Lyapunov exponent (LLE), detailed in Equation \ref{eq:LLE} in the Methods section, describes the layer's transient dynamics by setting $T=1$.

\begin{align}
\lambda[T]({\bf x}_i) &= \max_{\delta {\bf x}_i}{\frac{1}{T}\log{\frac{\parallel \delta {\bf x}_{i+T} \parallel}{\parallel \delta {\bf x}_i \parallel}}} \\
\lambda_{1}^{\rm LMLE}({\bf x}_i) &= \lambda[1]({\bf x}_i) \\
\lambda_{T}^{\rm FTMLE}({\bf x}_i) &= \lambda[T]({\bf x}_i) \\
\lambda_{\infty}^{\rm MLE} &= \lim_{T \to \infty} \lambda_{T}^{\rm FTMLE}({\bf x}_0)
\end{align}

\subsubsection{Multilayer Perceptron}
Comparable experiments evaluating the FTMLE in shallow MLPs have been conducted in prior research \cite{Misaki:Kondo:2021}. We expanded upon this analysis by tracking the FTMLE values of individual samples across each hidden layer. The MLP is trained for multi-class classification tasks. Specifically, we selected a two-dimensional input state space, enabling us to monitor the change in FTMLE values along the propagation through hidden layers. As depicted in Figure~\ref{fig:DNN-FTMLE}A, the outcomes indicated that the trained MLP overall exhibited negative FTMLE values, with the first three layers predominantly displaying positive values. By contrast, the later layers, which were responsible for finer-grained classification, tend to exhibit negative FTMLE values, which is indicative of a contraction behavior. In addition, the FTMLE heat-map shows the formation of the decision boundaries in hidden states, wherein broader linear boundaries progressively transformed into more specific non-linear boundaries. Regions with positive FTMLE values suggested an expansion behavior of features closely associated with the decision-making process. Additional experiments carried out on various datasets are provided in the Appendix \ref{sec:sup_ftmle_mlps}. A comparison of these results to the FTMLE analysis of untrained MLPs in Appendix \ref{sup:untrained_dnns} reveals a notable observation: the emergence of positive FTMLE values. This finding suggests that expansion properties are employed during training to facilitate classification via the separation of samples.

\subsubsection{Convolutional Neural Network}

We employed a pre-trained ResNet50 model \cite{he:2016:deep}, which had been trained on the ImageNet dataset \cite{deng:imagenet:2009}, and conducted our analyses using a sub-sampling from its validation set. As depicted in Figure~\ref{fig:DNN-FTMLE}B, our analysis of ResNet50 revealed an overall positive FTMLE distribution. Specifically, the initial max-pooling layer and the subsequent four bottleneck layers exhibited positive FTMLE values, implying that the expansion property was primarily harnessed for both low-level and high-level feature extraction purposes. Conversely, in the final average-pooling layer, where feature contraction occurred to generate predictions, we observed negative FTMLE values. In comparison to the untrained CNNs described in Appendix \ref{sup:untrained_dnns}, where certain layers displayed positive FTMLE values possibly due to the network's structure, we noticed that the training process enhances the expansion behavior. This enhancement was evidenced by the remarkably more positive post-training FTMLE distributions. These findings suggest that the network effectively utilized the expansion property during the feature extraction stage.

\subsubsection{Auto-encoder}
In addition to classification tasks, our study also examined image generation as a regression task. A pre-trained AE based on ResNet-18 was loaded on the CIFAR10 dataset \cite{krizhevsky:2009:learning} for this purpose. The task requires that output images retain the same dimensions as the input images while displaying feature compression in the encoder and decompression in the decoder. As depicted in Figure~\ref{fig:DNN-FTMLE}C, the overall network exhibited expansion property; however, when examined separately, the encoder and decoder clearly aligned with the task requirements by exhibiting contraction and expansion behaviors, respectively. When contrasted with the untrained AEs that displayed purely negative overall model FTMLE distributions, as shown in Appendix \ref{sup:untrained_dnns}, it is evident that the training process endowed the model with the expansion property, enabling it to effectively encode and decode input images.

\subsubsection{Bidirectional Encoder Representations from Transformers (BERT)}

BERT \cite{devlin:2018:bert} serves as a promising backbone network for learning feature representations in natural language processing tasks. For computation efficiency, we employed BERT-mini \cite{turc:2019:well}, a more compact variant of BERT equipped with four transformer encoder layers and an embedding size of 256. To delve into the internal dynamics of feature propagation across transformer layers in BERT-mini, we conducted our analysis using the IMDb benchmark \cite{maas:2011:learning}. 

As illustrated in  Figure~\ref{fig:DNN-FTMLE}D, the overall model and its sub-layers exhibited positive FTMLE values. Notably, each sub-layer manifested dominant expansion behavior, suggesting that the encoding process of language tokens necessitates the utilization of the expansion property for effective language representation learning. Compared to the evaluation of the untrained BERT-mini model detailed in Appendix \ref{sup:untrained_dnns}, the training process demonstrates a significant enhancement in the model's expansion behavior. Appendix \ref{sec:sup_bert_ftmle} provides a more in-depth analysis of the relationship between the FTMLE distribution and input token density. We therefore conclude that input sentences with richer content tokens (less padding tokens) are more likely to expand, thus preserving their information during layer propagation. By contrast, input sentences with fewer content tokens (more padding tokens) are more inclined to contract as they contain less information. \\

Our experiments revealed that the expansion property was consistently observed across various state-of-the-art DNNs. Furthermore, in contrast to the randomly initialized DNNs before training, which demonstrated either no or only weak expansion behavior, the DNNs post-training exhibited strong expansion behavior. This suggests that the training process equipped or reinforced the models with this capability for feature expansion. These observations lend strong support to the notion that expansion maps are utilized in DNNs to facilitate information propagation across layers.

\subsection{Exploiting Chaotic Systems as Computation Medium}

Driven by the above observation, we propose a novel framework that incorporates chaotic systems into neural networks as backbone models. To harness the intrinsic expansion property of chaotic systems, we simply introduced linear transformation layers before and after the dynamical system. These two linear layers can be tuned by leveraging back-propagation, while the intermediate system remains fixed to preserve its intrinsic dynamics. We chose a variety of chaotic systems, encompassing both discrete and continuous types, as the computational medium to show the generalization capability of our proposed method.

\subsubsection{Feed-Forward Echo-State Network (FFESN)}
\label{sec:ffesn}

\begin{figure*}[t!]
\centering
    \includegraphics[width=17.8cm]{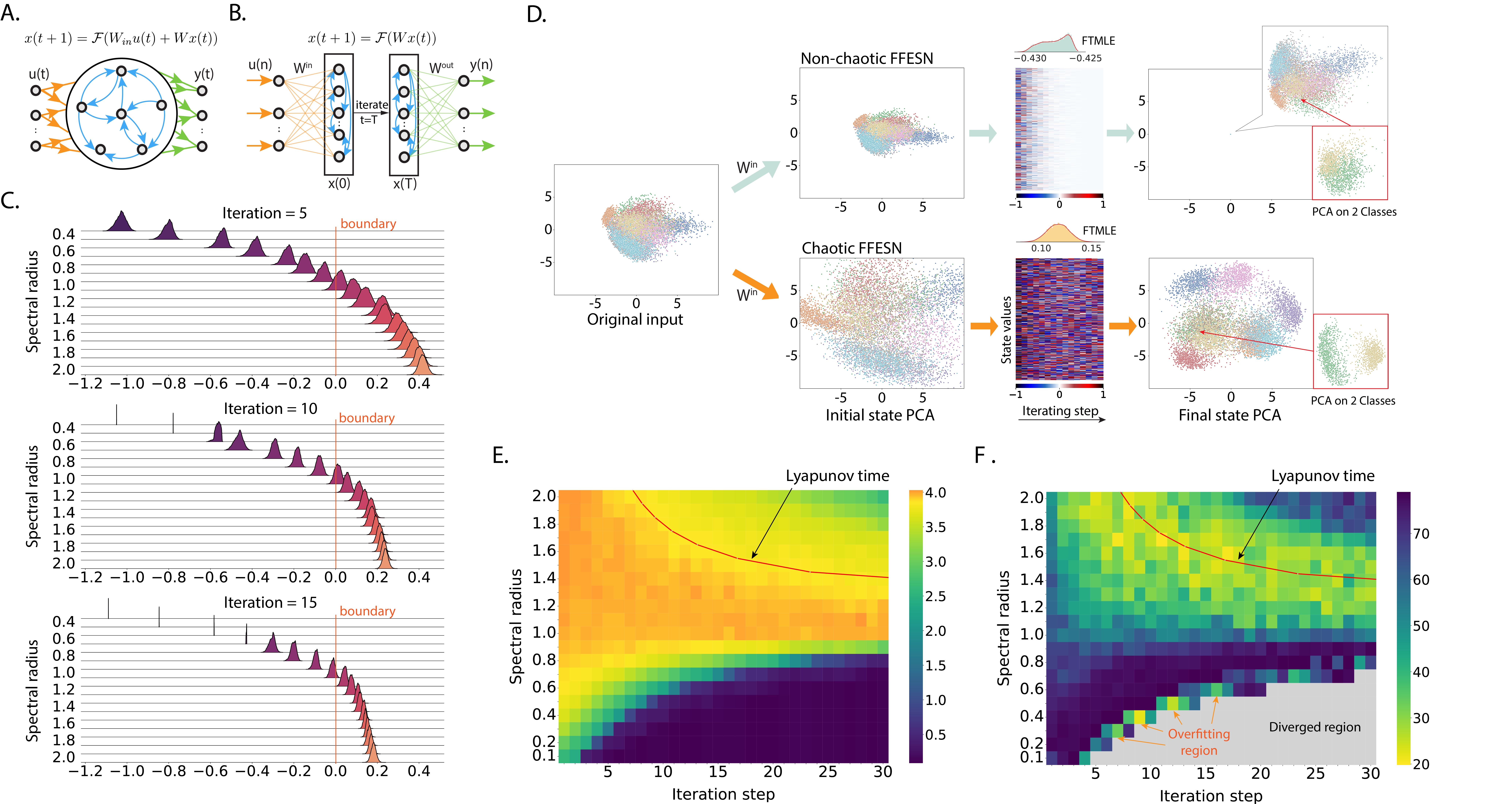}
\caption{The architecture and analysis of the FFESN. The training and evaluation are conducted on the MNIST dataset. \textbf{A.} Traditional ESN framework. \textbf{B.} The FFESN model takes a linearly transformed input as the initial state. The forthcoming state solely depends on the reservoir's present state. \textbf{C.} FTMLE analysis of the FFESN, considering a spectral radius $\rho$ varying from 0.3 to 2.0. Iteration steps $T$ chosen as 5, 10, and 15, corresponding to the upper, middle, and lower graphs, respectively. Each pinnacle on the ridge signifies an FTMLE distribution for the reservoir at each $T$. $\rho$ under 0.3, exhibiting more negative trends, are excluded. \textbf{D.} Comparison of classification capability between FFESNs with $\rho$ of 0.6 and 1.4 at $T=15$. This includes initial and final reservoir states and neuron state dynamics over time, with neuron numbers on the vertical axis, timestamps on the horizontal, and neuron values color-coded. FTMLE distributions are shown for $T=15$, with expansion behavior indicated by orange. PCA on final state images highlights differences in specific categories. \textbf{E.} The heat map reflects testing accuracy across $\rho$ and $T$ values, using $\left| \log(\epsilon) \right|$ for prediction error shaded in colors, with higher values indicating better accuracy. The red line shows the Lyapunov time, denoting the system's predictability time-frame \textbf{F.} The heat map displaying the FFESN convergence rate, based on epochs needed for optimal accuracy within a 5e-4 error margin, averaged over five trials. The gray area indicates performance akin to random predictions, and the orange arrows point to overfitting areas.}
\label{fig:CESN}
\end{figure*}

The echo-state network (ESN) emerges as a distinctive subcategory of RNNs and is distinguishable by its non-trainable recurrent layer \cite{jaeger:2004:harnessing}. Figure~\ref{fig:CESN}A illustrates the conventional network architecture of ESN designated for time-series data, channeling it through internal reservoirs. Base on the ESN framework, we introduce a novel feed-forward ESN (FFESN). This architecture eliminates the requirement for time-series input $u(t)$, instead receiving input data via a linear transformation layer that determines the initial state of the ESN iteration layer, as depicted in Figure~\ref{fig:CESN}B. In the context of the FFESN, the dynamics of the FFESN are measured through transient behaviors, given the restricted iteration step $T$. These dynamics are characterized by the spectral radius $\rho$, which is the maximum absolute value among the eigenvalues of the recurrent layer's internal weight matrix, predetermined during model initialization. In standard ESNs, chaotic tendencies appear when $\rho > 1$ \cite{sompolinsky:1988:chaos, verstraeten2007experimental}. For consistency in analysis, we employed FTMLE to analyze the recurrent layer, aligning our approach with earlier DNN analyses, to probe its transient expansion or contraction properties. Figure~\ref{fig:CESN}C illustrates the FTMLE assessment of FFESNs initialized with $\rho$ values spanning 0.3 to 2.0 and $T$ selected from 5, 10, and 15. Notably, the transient behaviors of FFESNs exhibited expansion maps when $\rho > 1$. When $T$ became larger, its corresponding FTMLE distribution exhibited values closer to the MLE steered by $\rho$.

To directly illustrate the state changes resulting from the internal dynamics of the recurrent layer, we evaluated the trajectories of reservoir states within the recurrent layer for both non-chaotic and chaotic FFESNs. As depicted in Figure~\ref{fig:CESN}D, the principal component analysis (PCA) diagram of the chaotic FFESN displayed a more distinct separation trajectory compared to the non-chaotic ESN. In the case of a non-chaotic network, its inherent contracting nature led to a significant contraction of the state space throughout the iteration procedure. This contraction resulted in the gradual vanishing of reservoir neuron values, which ultimately trended toward zero. By contrast, the chaotic network not only retained the variance but also amplified the clarity of separation after certain iterative steps, which demonstrates the usefulness of the expansion map induced by chaos. To further investigate the effects of $\rho$ and $T$, we computed the average testing accuracy and the convergent epoch as the number of epochs before reaching optimal accuracy. Figure~\ref{fig:CESN}E displays the average accuracy heat map;  the optimal accuracy was found in the chaotic region where $\rho > 1$. Additionally, as $T$ increased, the accuracy of the chaotic network tended to be preserved, in contrast to the non-chaotic network, for which the accuracy suddenly dropped after a few iterations. 

Moreover, the red line represents the Lyapunov time, a key time measure that quantifies how long it takes for small disturbances to expand across the full magnitude of the attractor, offering a predictive time scale for the predictability of the system's behavior. This serves as an insightful indicator, identifying the configuration of $T$ for which the network achieves its optimal accuracy region. Figure~\ref{fig:CESN}F displays the average epoch at which convergence occurs during training. We observed that chaotic FFESNs tended to converge more rapidly when $T$ aligned with its associated Lyapunov time compared to other regions. This quicker convergence can likely be attributed to more effective weight optimization for linear layers, induced by chaotic iteration layers, throughout the training process. Additionally, with a relatively large $T$, the network suffered from overfitting, resulting in lower accuracy or even divergence. An evaluation of both optimal accuracy and convergence speed showed that the chaotic network exhibited a marked superiority over its non-chaotic counterpart. As shown in Appendix \ref{sec:sup_ffesn_fmnist}, we conducted an evaluation of FFESN using the Fashion-MNIST dataset and observed similar trends concerning the region of optimal accuracy and convergence speed. Furthermore, to explore the impact of chaos on convergence speed in a general case, we evaluated the convergence speed of MLPs initialized with weights having various spectral radius in Appendix \ref{sec:sup_mlp_init}. We discovered that chaos plays an active role in enhancing the speed of convergence.

\subsubsection{Lorenz 96}

\begin{figure*}[t!]
\centering
    \includegraphics[width=17.8cm]{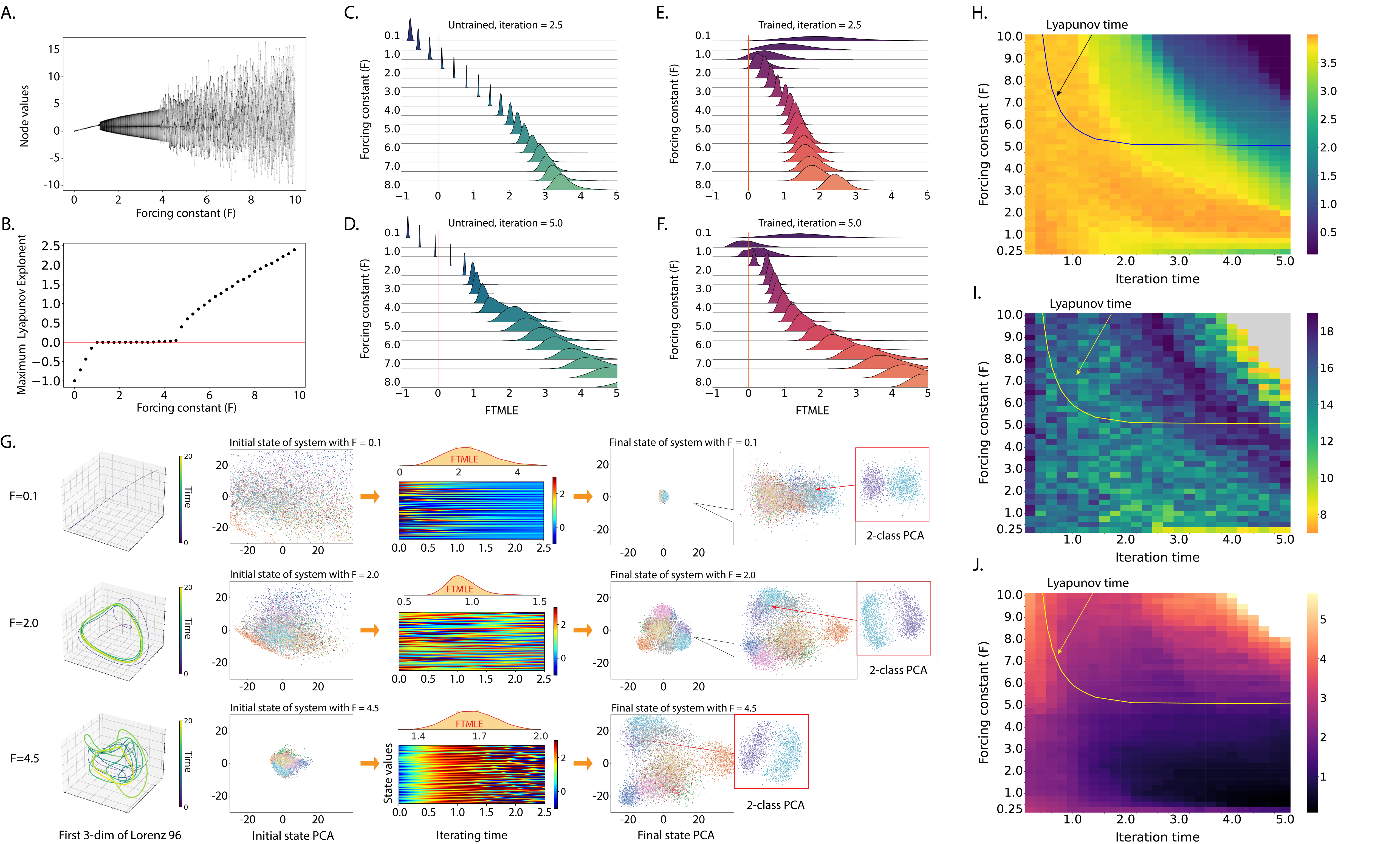}
\caption{Exploration of Lorenz 96 system performance on the MNIST task.
\textbf{A.} Global bifurcation diagram depicting the system behavior with random input over an iteration time $T=20$.
\textbf{B.} The MLE identifies global dynamics of fixed-point convergence, periodic orbits, and chaos.
\textbf{C--D.} Illustration of the FTMLE spectra for the untrained Lorenz 96 models.
\textbf{E--F.} Distribution of the FTMLE for the trained Lorenz 96 models, particularly highlighting the divergent transient dynamics observed at smaller $F$ values (e.g. less than 2) compared to its untrained counterpart.
\textbf{G.} PCA projections of initial, final states with FTMLE indicators. 3D trajectory diagrams show state evolution in Lorenz 96 system's first three dimensions, spanning $T$ from 0 to 20. PCA results use $T=2.5$ particularly. The horizontal axis of internal state dynamics represents iteration time, and the vertical axis shows state values. The color gradient indicates neuron value.
\textbf{H--J.} Heat maps illustrating classification accuracy, training epoch, and averaged FTMLE across varied $F$ and $T$. The accuracy map \textbf{H} has a color spectrum indicating $\left| \log(\epsilon) \right|$, where $\epsilon$ is the prediction error. The convergence metric \textbf{I} illustrates the number of training epochs necessary to achieve optimal accuracy, considering an error tolerance of $5e-5$. The  FTMLE map \textbf{J} shows averaged FTMLE over the testing set. }
\label{fig:Lorenz96}
\end{figure*}

We extended our framework to continuous-time internal layers to evaluate the effectiveness of real-world chaos. Here, we employ the Lorenz 96 \cite{lorenz:1996:predictability}, a model of the atmosphere that shows high-dimensional spatiotemporal chaos with continuous time, with trainable linear $W_{\rm in}$ and $W_{\rm out}$ added before and after the dynamical system, as formulated by the following equation:

\begin{eqnarray}
{\bf x}(0) &=& W_{\rm in} {\bf u}(n)~, \\
{\bf x}(t) &=& {\bf x}(0) + \int_{0}^t {\dot{{\bf x}}(s)ds} \label{continuousDNN}~,~\text{and} \\
{\bf y}(n) &=& W_{\rm out} [ {\bf x}(T)^{\top} ; 1]^{\top}~,
\end{eqnarray}
where $T \in {\mathbb{R}}$ represents iteration time, and $x \in \mathbb{R}^{\rm 500}$ is the state of the system. As depicted in Figure~\ref{fig:Lorenz96}A--B, the Lorenz 96 system's global dynamics reveal three salient regimes: fixed-point convergence, periodic oscillations, and chaotic behavior, each dictated by varying values of the forcing term ($F$). The MLE diagram elucidates these dynamics in relation to $F$. Notably, our experimental focus centered on harnessing the transient dynamics of the Lorenz 96 system for small $T$, which was assessed through FTMLE. Instead of applying conventional back-propagation to discrete internal layers, we employed the {\it adjoint sensitivity method} for error propagation within continuous-time internal systems \cite{pontryagin:1987:mathematical, chen:2018:advances}. 

In Figure~\ref{fig:Lorenz96}C--F, a comparative analysis between the FTMLE spectra of both trained and untrained systems across $T=2.5$ and $5.0$ is presented. The untrained system demonstrated a monotonic trend: with a rise in $F$, the FTMLE values tend to increase. By contrast, the trained system exhibits a deviating pattern. Specifically, at lower $F$ values, there is a positive FTMLE distribution. This behavior can be attributed to the training process, which allows the system to identify an optimal initial state by adjusting $W_{\rm in}$, thereby suppressing the system's natural tendency to contract. Additionally, a consistent observation across all systems was the predominance of positive FTMLE distributions, hinting at the utilization of an expansion map during propagation.

Further insights are provided in Figure~\ref{fig:Lorenz96}G, which highlights three representative systems corresponding to $F=0.1$, $2.0$, and $4.5$. The depicted 3D trajectory diagrams illuminate the longer trajectory ($T$ up to 20) for the initial three dimensions of the Lorenz 96 system, each visualizing a unique global dynamic. The subsequent state analysis was conducted at $T=2.5$ to assess transient dynamics. From the PCA visualizations, the contrasts between the initial and final states of the iterative internal system were evident. Specifically, at $F=0.1$, the initial state appeared markedly expanded, corroborating our hypothesis regarding the expansive resistance to contraction. This initially high-variance state underwent significant contraction throughout the iteration. State dynamics diagrams further illustrate the evolution of state values throughout the iterating period. A salient observation across all state dynamics was the presence of positive FTMLE values, again reaffirming the system's inclination toward expansion mapping. For small $F$ values, such as $F=0.1$, even when the feature space generally contracted, the distinct category clusters retained discernible boundaries, as illustrated in the two-class PCA visualization (PCA applied to two selected categories only).

Figure~\ref{fig:Lorenz96}H shows the averaged accuracy heat map, with higher values (lighter colors) indicating higher accuracy. Notably, the regions of optimal accuracy, as represented by the luminous yellow hues, predominantly lie in regions that include (i) $F < 2$ and $T < 1$, and (ii) $F \in (1, 2.5)$ and $T > 2.5$, where higher $F$ with relative small $T$ still yields good performance. Figure~\ref{fig:Lorenz96}I shows the results of recording the average number of epochs required to reach optimal accuracy. Disregarding the divergence region and the early overfitting region (low numbers of epochs and low accuracy), we observed the optimal region for fast convergence overlaps with the optimal accuracy region. Notably, as shown in Figure~\ref{fig:Lorenz96}J, all areas of the averaged FTMLE heat map display positive FTMLE values. The optimal region, distinguished by higher accuracy and faster convergence speed, exhibits relatively lower FTMLE values, yet still demonstrates expansion behavior. Furthermore, in contrast to FFESN in our main content, the Lyapunov time does not serve as an indicator for the region of optimal performance.

\subsubsection{Coupled Spin-Torque Oscillators}

\begin{figure*}[t!]
\centering
\includegraphics[width=17.8cm]{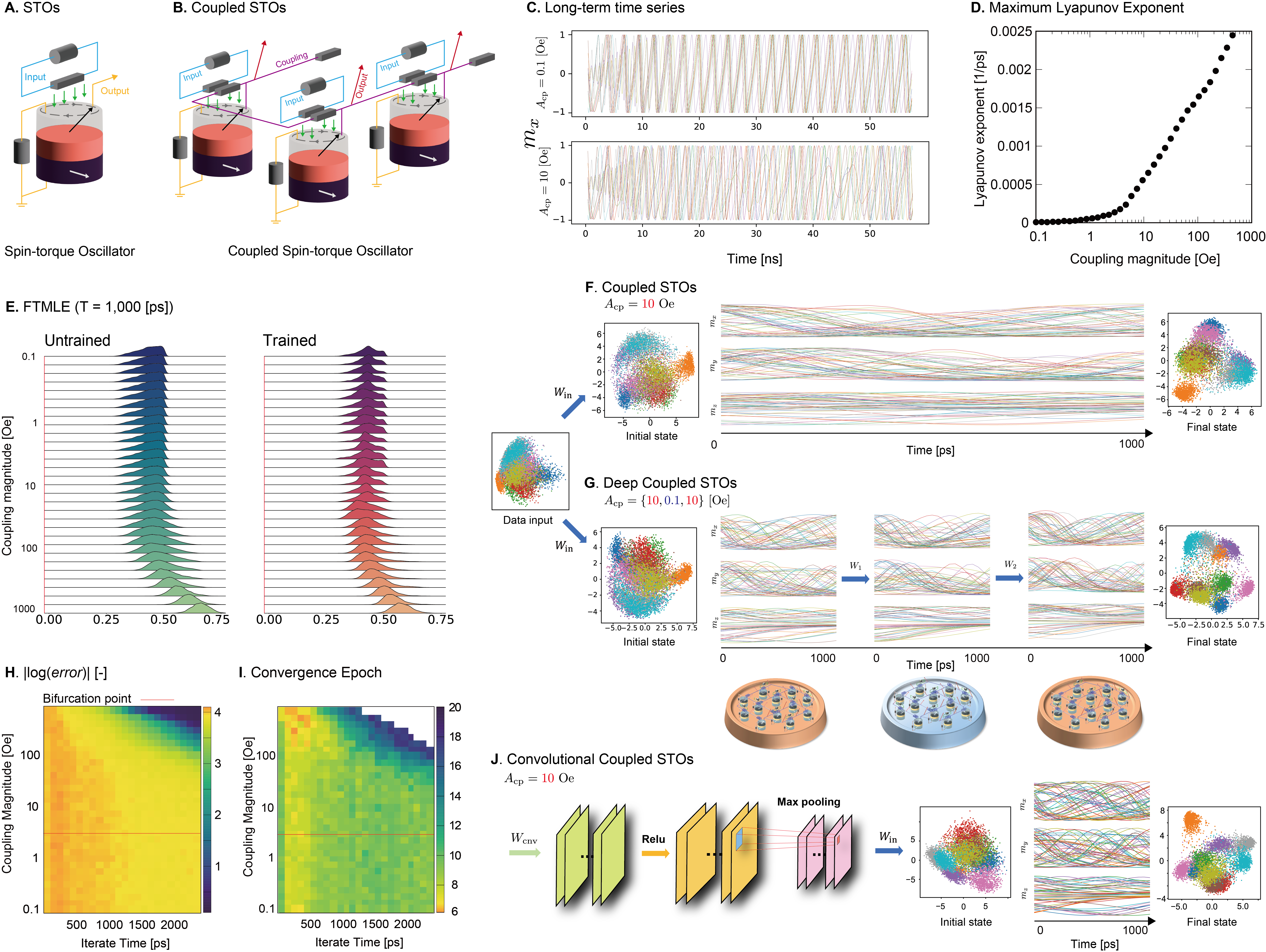}
\caption{Dynamical analyses for coupled STOs and solving MNIST task using coupled STOs. 
 \textbf{A--B.} The schematic diagrams of STOs and coupled STOs.
 \textbf{C.} Typical long-term dynamics of periodic and chaotic coupled STOs with random initial states.
 \textbf{D.} MLEs of coupled STOs through coupling magnitude $A_{\rm cp}$.
 \textbf{E.} FTMLE distributions for coupled STOs for MNIST input data through coupling magnitude $A_{\rm cp}$. The left and right graphs show the FTMLE distributions of untrained and trained coupled STOs, respectively.
 \textbf{F.} The architecture using coupled STOs as an internal layer.
 \textbf{G.} The deep architecture using three coupled STOs as internal layers.
 \textbf{H.} The MNIST performance heat map in terms of iteration time and coupling magnitude. The color represents the $\left| \log(\epsilon) \right|$.
 \textbf{I.} The MNIST training speed heat map through the iteration time and coupling magnitude. The color represents the convergence epoch.
 \textbf{J.} The architecture using three coupled STOs with a convolutional input layer. The scatter plots illustrate MNIST test data in the first and second principal component space. The time series are MNIST data transformations in computational mediums.
 }
\label{fig:spin}
\end{figure*}

We implemented our approach on spin-torque oscillators (STOs), which are a type of spintronics device, as depicted in Figure~\ref{fig:spin}A. STOs are considered a promising option for neuromorphic computing devices due to their small size, high-speed dynamics, and high energy efficiency \cite{torrejon:2017:neuromorphic,tsunegi:2019:physical,markovic:2020:physics,grollier:2020:neuromorphic}. Typically, magnetization in STOs exhibits limit cycle oscillation dynamics, although there are reports of chaotic dynamics influenced by input \cite{yang:2007:chaotic, yamaguchi:2019:synchronization, akashi:2020:input}, feedback \cite{taniguchi:2019:chaos, kamimaki:2021:chaos}, and coupling \cite{akashi:2022:coupled}. However, previous research has not effectively utilized chaotic dynamics for information processing. Thus, we concentrated on leveraging high-dimensional chaotic dynamics in coupled STOs, simulating these dynamics by solving the Landau-Lifshitz-Gilbert equation \cite{akashi:2022:coupled}. Details on the device configurations and simulation settings are provided in the Method section.

The behavior of magnetization in coupled STOs exhibits a bifurcation between periodic and chaotic dynamics, influenced by the coupling magnitude $A_{\rm cp}$. Figure~\ref{fig:spin}C illustrates a typical extended time series of the magnetizations in these coupled STOs. Specifically, the magnetization in coupled STOs with $A_{\rm cp} = 0.1$ and $10$ Oe showed periodic and chaotic dynamics, respectively. Figure~\ref{fig:spin}D presents the MLE of the coupled STOs across different values of $A_{\rm cp}$. We observed that the MLE shifted from 0 to positive, signifying a transition from periodic to chaotic dynamics, as $A_{\rm cp}$ increased.

In our proposed framework, we utilized the dynamics of coupled STOs as an internal backbone model, with the mathematical formulations provided in the Methods section. For the MNIST dataset, the input data were converted into the states of coupled STOs. These states represent the magnetization of each STO in a three-dimensional form ${\bf m}_k = ( m_{k, x} m_{k, y} m_{k, z})^{\top}$, based on the input weight. Our network comprised 200 coupled STOs, resulting in a total of 600 computational nodes, corresponding to the three dimensions of each STO's magnetization.

Our analysis encompassed two scenarios involving input data from both untrained (randomly initialized) and trained input weights. Despite the occurrence of bifurcation in the long-term dynamics, the FTMLE consistently remained positive in both scenarios, detailed in Appendix \ref{sec:sup_csto_ftmle}. This consistency indicates that the transient dynamics were chaotic. Figure~\ref{fig:spin}F--G illustrate the flow of MNIST data transformed by a specific example of chaotic coupled STOs and a chaotic deep architecture. Additionally, we present heat maps depicting the performance and convergent speed of coupled STOs in Figure~\ref{fig:spin}H--I. Both the limit cycle and chaotic regions exhibited good performance with an appropriate iteration time, which was around 300 ps. However, it is noteworthy that the peak in training speed occurred in the intensely chaotic region ($A_{\rm cp} > 100$ Oe). This performance is on par with state-of-the-art neuromorphic spin-torque devices for the MNIST task \cite{leroux:2022:convolutional}. Such a pattern aligns with observations made in both the FFESN and Lorenz 96 systems. For the Lyapunov time, we found no correlation with either performance or convergence speed, and thus, it is not presented here. 

For a direct comparison between discrete-time and continuous-time dynamical systems, as well as conventional DNNs, we conducted experiments using FFESN, Lorenz 96, and coupled STOs under specific settings. These were compared with a linear regression model, an MLP with one hidden layer, and a CNN with one convolutional layer and one hidden fully connected layer. To ensure a fair comparison, the number of hidden neurons in the MLP, FFESN, and Lorenz 96 models was identical. Furthermore, we applied a convolutional layer to the coupled STOs, as shown in Figure~\ref{fig:spin}J, with a read-in layer that is equivalent to the baseline convolutional layer. The highest accuracy and the average accuracy across five trials for each model were recorded, as presented in Table~\ref{table:mnist_accuracy}. The specific settings for each model are detailed in Appendix \ref{sec:sup_framework_eq}. We observed that FFESN and Lorenz 96 systems showed superior performance against MLP, and coupled STOs with convolutional input layer outperformed the conventional CNN as well, presenting state-of-the-art performance in neuromorphic computing.

\begin{table}[t!]
\centering
\setlength{\tabcolsep}{0.7pt}
\caption{Accuracy comparison of the proposed framework on the MNIST dataset, showing the highest and average accuracy across five trials for all models.}
\label{table:mnist_accuracy}
\scriptsize
\begin{tabular}{lcc}
\toprule
System & High.(\%) & Avg.(\%) \\
\midrule
Linear Regression & 92.52 & 92.48 \\
MLP ($N=500$) & 97.53 & 97.45 \\
FFESN ($\rho=0.9$, $N=500$, $T=1$) & 98.16 & 98.08 \\
FFESN ($\rho=1.0$, $N=500$, $T=1$) & 98.26 & 98.12 \\
FFESN ($\rho=1.8$, $N=500$, $T=2$) & 98.34 & 98.23 \\
Lorenz 96 ($F=0.5$, $N=500$, $T=0.4$) & 98.29 & 98.17 \\
Lorenz 96 ($F=4.75$, $N=500$, $T=0.2$) & 98.20  & 98.02 \\
\midrule
CNN ($N=600$) & 98.77 & 98.69 \\
CSTOs ($N=600$, $A_{\rm cp}=17.8$ Oe, $T=300$ ps) & 98.43  & 98.33 \\
Deep CSTOs ($A_{\rm cp}=\{10, 0.1, 10\}$, $T=100$ ps) & 98.56  & 98.33 \\
CNN + CSTOs ($N=600$, ${\rm cp}=745$ Oe, $T=200$ ps) & \textbf{99.05}  & \textbf{99.00} \\
\bottomrule
\end{tabular}
\text{MLP and CNN all have only one hidden layer with $N$ neurons.}
\end{table}

In addition to the superiority of accuracy and convergence speed, our coupled STOs system also demonstrated robustness to noise. In Appendix \ref{sec:sup_csto_noise}, we exposed the system to two types of noise, and it showed consistent performance, regardless of the noise level.

\section{Discussion}

In the present investigation, we delved into the transient dynamics of state-of-the-art DNNs across diverse domains through the lens of dynamical systems, employing layer-wise FTMLE analyses. Our findings underscore the pervasive use of expansion maps within DNNs to facilitate information processing, such as the encoding of input features. Furthermore, the assessment of FTMLE presents a novel approach to interpreting DNNs within the framework of dynamical systems, proffering a fresh perspective on the interpretability of DNNs.

Motivated by the prevalent employment of expansion maps in DNNs, we introduce a straightforward framework that exploits chaotic dynamics as a computational medium. We focus on three distinct systems: (1) the discrete-time FFESN, (2) the continuous-time Lorenz 96, and (3) a neuromorphic system known as the STOs. Our experimental design encompasses a range of settings, representing various global dynamics. A subsequent deep dive into the transient dynamics of these systems revealed certain deviations from global dynamics, which are particularly pronounced in continuous-time dynamical systems.

Through the experiments conducted on these varied systems, we discovered that chaotic dynamics can indeed be utilized for information processing, with our empirical tasks centered on classification. A salient observation was the role of expansion maps in these systems, as evidenced by dominant positive FTMLE distributions, which consistently led to optimal performance. This concurrence of expansion behavior with optimal classification accuracy underscores the potential of chaotic dynamics for practical applications. For example, our successful integration of STOs illustrates the viability of incorporating real-world chaotic systems. These systems naturally progress over time and obviate the need for state computation, thereby opening new paths for building energy-efficient neuromorphic systems.

However, our study faces certain limitations. Primary among these is the fact that our framework predominantly focuses on basic image classification tasks, ones in which DNNs demonstrate marked proficiency. It remains a considerable challenge to attain superior performance on more difficult tasks compared to state-of-the-art DNNs. A barrier to this challenge is the inherent difficulty in augmenting the depth of integrated systems. Nonetheless, our research unveils the usefulness of chaotic dynamics in which expansion maps are employed. This insight presents a novel avenue for the incorporation of chaotic components in the design of DNN architectures in forthcoming research. A secondary concern pertains to the practical integration of STOs in real-world applications. Specifically, managing the initial state, represented by the electron’s spin angle, is challenging to control with precision in reality. However, it is plausible to explore methods for controlling the state via input signals, drawing on the principles of input synchronization. Our empirical approach to employing chaotic dynamics in neuromorphic computing highlights the untapped potential of chaotic systems—once considered inconsequential. This revelation not only showcases the utility of such systems but also suggests that other physical reservoirs, with STOs being just one example, might offer even greater implementation ease for this scheme. These insights pave the way for exploring more complex integrations in future neuromorphic architectures, broadening the scope beyond our current examples.

\section{Materials and Methods}

\subsection{Finite-Time Maximum Lyapunov Exponent (FTMLE)}
In dynamical system theory, the MLE measures the average rate of divergence or convergence among neighboring trajectories in a system's state space. By setting $T=\infty$ and the initial state as ${\bf x}_0$ in Equation \ref{eq:general_LE}, we derive the MLE as follows:

\begin{align}
    \lambda_{\infty}^{\rm MLE} &= \lim_{T \to \infty} \lambda[T]({\bf x}_0) \label{eq:MLE} ~.
\end{align}

The MLE serves as a measure of the stationary dynamics of the model. Conversely, when $T=1$, we can determine the LLE as a measure of the transient dynamics of a single layer in the term:

\begin{align}
   \lambda_{1}^{\rm LLE}({\bf x}_i) &= \lambda[1]({\bf x}_i) ~. \label{eq:LLE}
\end{align}

Unlike the MLE, the FTMLE is especially well-suited for our experiments as it adeptly captures the transient dynamics in both discrete-time and continuous-time systems. It is also notable for its computational efficiency. In contrast to the LLE, which necessitates a laborious layer-by-layer calculation, the FTMLE provides an effective assessment over a finite number of layers without compromising accuracy.

In our experiments, we evaluated the FTMLE both from the perspective of the overall model and on a layer-wise basis. For the overall model's FTMLE assessment, we define ${\bf x}_0$ as the input data entering the model, and ${\bf x}_n$ represents the output of the model after passing through $n$ layers. Similarly, for the layer-wise FTMLE analysis, ${\bf x}_i$ is set as the input to a specific sub-layer that consists $n$ layers, and ${\bf x}_{i+n}$ is the output from the target sub-layer. Since $\delta {\bf x}_{i+1}$ can be computed as:
\begin{equation}
\delta {\bf x}_{i+1} = \frac{\partial F_{i}}{\partial {\bf x}_{i}} \delta {\bf x}_{i} ~,
\end{equation}

\noindent by the chain rule, the FTMLE, defined in Equation~\ref{eq:FTMLE}, can be derived as:

\begin{align}
    \lambda_{n}^{\rm FTMLE}({\bf x}_0) &= \max_{\mathcal{J}\in\mathbb{R}^{{\rm D_0} \!\times\! {\rm D_n}}} \frac{1}{n} \log{\parallel \mathcal{J}_{n} \mathcal{V} \parallel} ~,
\end{align}

\noindent where $\mathcal{J}_{n}$ denotes the Jacobian matrix obtained after propagation through $n$ layers, and $\mathcal{V}$ symbolizes $\delta x_{0}$, which characterizes the perturbation exerted on the initial state $x_{0}$. Subsequently, $\parallel \mathcal{J}_{n} \mathcal{V} \parallel$ can be derived as:

\begin{align}
 \parallel \mathcal{J}_{n} \mathcal{V} \parallel &= \sqrt{\mathcal{V}^T \mathcal{J}_{n}^T \mathcal{J}_{n} \mathcal{V}} \nonumber \\
 &= \sqrt{\mathcal{V}^T P_{n}^T \Lambda_{n} P_{n} \mathcal{V}} \nonumber \\
 &= \sqrt{ \mathcal{V}^T P_{n}^T
  \left[ \begin{smallmatrix}
  {\sigma_1}^2 & & \\
  & \ddots & \\
  & & {\sigma_D}^2
  \end{smallmatrix} \right]_{n}
 P_{n} \mathcal{V}} ~.
 \label{eq:VP_SIG}
\end{align}

\noindent Let $\mathcal{U}$ denotes $P_{n} \mathcal{V}$; then Equation \ref{eq:VP_SIG} can be derived as

\begin{align}
\parallel \mathcal{J}_{n} \mathcal{V} \parallel
 &= \sqrt{ \mathcal{U}^T
  \left[ \begin{smallmatrix}
  \sigma_1 & & \\
  & \ddots & \\
  & & \sigma_D
  \end{smallmatrix} \right]_{n}
  \left[ \begin{smallmatrix}
  \sigma_1 & & \\
  & \ddots & \\
  & & \sigma_D
  \end{smallmatrix} \right]_{n}
 \mathcal{U} } \nonumber \\
 &= \parallel
  \Sigma_{n} \mathcal{U}
  \parallel.
\end{align}

\noindent When $\mathcal{U}$ becomes the eigenvector of $\mathcal{J}_{n}^T \mathcal{J}_{n}$ with respect to $\sigma_1$, $\parallel \mathcal{J}_{n} \mathcal{V} \parallel$ is maximized:

\begin{align}
\lambda_{n}^{\rm FTMLE}({\bf x}_0) &= \max_{\mathcal{J} \in \mathbb{R}^{{\rm D_0} \!\times\! {\rm D_n}}} \frac{1}{n} \log{\parallel \mathcal{J}_{n} \mathcal{V} \parallel} \nonumber \\
&= \frac{1}{n} \log{\sigma_{n}^1} ~,
\end{align}

\noindent where $\sigma_{n}^{1}$ is the maximum singular value of $\mathcal{J}_{n}$. To compute $\mathcal{J}_{n}$, we utilized {\it pytorch.autograd} for efficient gradient calculations. Subsequently, we employed SVD decomposition on the obtained Jacobian matrix. In scenarios in which the input and output spaces are exceedingly large, as is the case with CNNs, traditional SVD computations can become exponentially expensive. In response, we adopted a truncated randomized SVD decomposition approach, wherein we calculated a limited number of five candidate singular values, denoted as $m$, and selected the maximum one. In comparison with the method suggested in previous work \cite{Misaki:Kondo:2021}, our approach yields more accurate FTMLE values, with the trade-off of requiring more computation efforts.

\subsection{Feed-Forward Echo-State Network (FFESN)}
\label{result:FFESN}
The ESN is a RNN variant, falling within the broader category of reservoir computing methods; it is uniquely tailored for tasks involving temporal sequences. Distinctively, ESNs simplify learning by maintaining fixed weights in the recurrent reservoir layer while adjusting solely the output weights during training.

Based on the ESN, we propose the FFESN, in which the network's initial state is derived directly from the input data. This alternative architecture no longer takes time-series input; instead, similar to a MLP, the current state is solely dependent on the previous state:

\begin{align}
x(0) &= W_{\rm in}u ~, \\
x(i+1) &= f(W'x(i)) ~,~\text{and} \\
y &= W_{\rm out}x(n) ~.
\end{align}

\noindent Here, $u$ represents the input data, $y$ is the output, $W_{\rm in}$ and $W_{\rm out}$ denote two trainable linear layers, $W'$ represents the fixed intrinsic weight, and $x(i)$ represents the state at iteration $i$. Within this FFESN context, the ESP is no longer applicable.

In our experiments, we set the number of internal neurons of the FFESN to 500. For $W'$, the connectivity density is set to 0.5, and the weight is initialized via spectral radius ranging from 0.1 to 2.0. During the training of the FFESN model, we employed stochastic gradient descent (SGD) with a learning rate of 5e-3, momentum of 0.9, and batch size of 64.

\subsection{Lorenz 96}
The Lorenz 96 model serves as a critical benchmark for understanding the dynamical behavior of complex systems. Characterized by a series of ordinary differential equations, this model effectively captures the essence of nonlinear, chaotic dynamical systems. Mathematically, the model is defined by this equation:

\begin{equation}
\frac{dx_i}{dt} = (x_{i+1} - x_{i-2})x_{i-1} - x_{i} + F ~.
\end{equation}

In this system, the variable $x_{i}$ denotes the state at the $i^{th}$ neuron (dimension) of the system, and the state is defined as ${\bf x}(t) = (x_{1}(t) \cdots x_{N}(t))^{\top}$. The parameter $F$, referred to as the forcing term, is a constant that drives the internal dynamics of the system. To ensure continuity and cyclical behavior intrinsic to many physical phenomena, the system state variables are arranged in a periodic fashion. Specifically, we consider that the system is composed of $N$ sites, with $N \geq 4$, and the system variables are arranged cyclically such that $x_{-1}=x_{N-1}$, $x_{0}=x_{N}$, and $x_{N+1}=x_{1}$. This condition elegantly emulates a circular system in which each site communicates with its neighboring sites in a cyclic manner, ensuring that the system is devoid of boundary conditions.

The inherent flexibility of the Lorenz 96 model allows it to be scaled to higher-dimensional spaces, thereby adapting it to more complex systems or networks. In the context of network models, such scaling can be utilized to accurately match the state space of the underlying or hidden state of the networks. This scaling, while maintaining the fundamental properties of the Lorenz 96 system, allows the exploration of complex network behaviors in a robust and consistent framework, building upon the established principles of dynamical systems theory.

\subsection{Coupled Spin-torque Oscillators}

With regard to neuromorphic computing, parallel conclusions have been drawn, suggesting that transient chaotic dynamics can stimulate the system to escape local minima, facilitating convergence to a global minimum or the optimal solution \cite{kumar:2017:chaotic}. Furthermore, the amalgamation of periodic and chaotic oscillations within a nanoscale third-order circuit element has been shown to produce superior statistics in error minimization \cite{kumar:2020:third}.

The magnetic tunnel junction (MTJ) is a spintronics device that consists of two ferromagnetic metals and a nonmagnetic spacer.
Two magnetic layers are classified as the free and reference layers. The magnetization of the reference layer is fixed.
The magnetization of the free layer exhibits various dynamics, such as limit cycle oscillation, converging to a fixed point, and chaos, after the injection of magnetization. We can detect the magnetization of the free layer as an output current via the tunnel magnetoresistance effect. In this study, we made STOs a high-dimensional system by coupling them together by means of magnetic fields to exploit them as a computational network~\cite{akashi:2022:coupled}. We interconnected STOs by magnetic fields, which are derived from the output currents of other STOs through Ampere's law.

We exploited the dynamics of magnetization as a physical deep neural network. The dynamics of magnetization ${\bf m}_k = (m_{k,x}, m_{k,y}, m_{k,z})^{\top} \in \mathbb{R}^3 $ of coupled STOs are described by the following LLG equation:

\begin{eqnarray}
\label{eq:llg}
\frac{d{\bf m}_k}{dt} &=& - \gamma {\bf m}_k \times \left[{\bf H}_{k} + {\bf H}^{{\rm cp}}_{k} \right] \nonumber \\
&& - \gamma H_{s,k} {\bf m}_k \times ({\bf p}\times {\bf m}_k) \nonumber \\
&& + \alpha {\bf m}_k \times \frac{d{\bf m}_k}{dt} ~\text{and} \\
H_{s,k} &=& \frac{\hslash \eta I}{2e(1+\lambda {\bf m}_{k}\cdot{\bf p}) M V} ~,
\end{eqnarray}

\noindent where the magnetic field 
\begin{eqnarray}
{\bf H}_k = \left[H_{{\rm appl}} + (H_{{\rm K}}-4 \pi M)m_{k,z} \right]{\bf e}_z
\end{eqnarray}

comprises an applied field $H_{\rm appl}$.
The coupling magnetic field ${\bf H}^{{\rm cp}}_{k}$ is described by the following equation:
\begin{eqnarray}
\label{eq:llg_coupling}
{\bf H}^{{\rm cp}}_{k} &=& A_{{\rm cp}}\sum_{i}^{N}{w^{{\rm cp}}_{k,i} m_{i,x} {\bf e}_{x}} ~, 
\end{eqnarray}
\noindent where $A_{{\rm cp}}$ is the coupling magnitude, and ${\bf e}_{x}$ is the $x$ directional unit vector. 
$W_{\rm cp} = (w^{{\rm cp}}_{k,i}) \in {\mathbb{R}^{N\times N}}$ is the internal couping weight, of which the components are random variables drawn from the uniform distribution within the interval $[-1, 1]$.
We scaled $W_{\rm cp}$ so that its spectral radius is 1.
Other parameters and the values in Equation.~\ref{eq:llg} are provided in Appendix \ref{sec:sup_csto_parameters}.

Further, ${\bf m}_k = (m_{k,x}, m_{k,y}, m_{k,z})^{\top}$ is restricted to the unit sphere surface.
In other words, $\|{\bf m}_k \|| = 1$ always holds.
Therefore, we applied a normalizing function $f_{\rm norm}({\bf m_k}) = {\bf m_k} / \|{\bf m}_k \|$ after injecting input data into coupled STOs.

\begin{acknowledgments}
The authors would like to acknowledge Dr. Sumito Tsunegi and Dr. Tomohiro Taniguchi for their helpful suggestions on the physical settings of the spin-torque oscillators used for neuromorphic simulation.\\\\
\noindent \textbf{Funding:} This study was partially supported by a JST CREST Grant, Number JPMJCR2014, Japan. \\\\
\noindent \textbf{Author Contributions} N.A. and K.N. led the research program and proposed the initial concept of utilizing chaotic dynamics for computation. S.L., H.Q., N.A., and K.N. developed the algorithms. S.L., H.Q., and N.A. performed the experiments, prepared the results, and drew the diagrams. S.L. and N.A. wrote the main content of the paper. S.L., N.A., and H.Q. prepared the supplementary information. Y.K. provided useful suggestions on the motivation and experiment settings. K.N. for the organization of the overall paper and approved the submission.\\\\
\noindent \textbf{Competing Interests} The authors declare no competing interests.\\\\
\noindent \textbf{Data and Materials Availability:} The code supporting the findings of this study is available from the corresponding author upon request.

\end{acknowledgments}

\bibliography{apssamp}

\begin{thebibliography}{74}%
\makeatletter
\providecommand \@ifxundefined [1]{%
 \@ifx{#1\undefined}
}%
\providecommand \@ifnum [1]{%
 \ifnum #1\expandafter \@firstoftwo
 \else \expandafter \@secondoftwo
 \fi
}%
\providecommand \@ifx [1]{%
 \ifx #1\expandafter \@firstoftwo
 \else \expandafter \@secondoftwo
 \fi
}%
\providecommand \natexlab [1]{#1}%
\providecommand \enquote  [1]{``#1''}%
\providecommand \bibnamefont  [1]{#1}%
\providecommand \bibfnamefont [1]{#1}%
\providecommand \citenamefont [1]{#1}%
\providecommand \href@noop [0]{\@secondoftwo}%
\providecommand \href [0]{\begingroup \@sanitize@url \@href}%
\providecommand \@href[1]{\@@startlink{#1}\@@href}%
\providecommand \@@href[1]{\endgroup#1\@@endlink}%
\providecommand \@sanitize@url [0]{\catcode `\\12\catcode `\$12\catcode `\&12\catcode `\#12\catcode `\^12\catcode `\_12\catcode `\%12\relax}%
\providecommand \@@startlink[1]{}%
\providecommand \@@endlink[0]{}%
\providecommand \url  [0]{\begingroup\@sanitize@url \@url }%
\providecommand \@url [1]{\endgroup\@href {#1}{\urlprefix }}%
\providecommand \urlprefix  [0]{URL }%
\providecommand \Eprint [0]{\href }%
\providecommand \doibase [0]{https://doi.org/}%
\providecommand \selectlanguage [0]{\@gobble}%
\providecommand \bibinfo  [0]{\@secondoftwo}%
\providecommand \bibfield  [0]{\@secondoftwo}%
\providecommand \translation [1]{[#1]}%
\providecommand \BibitemOpen [0]{}%
\providecommand \bibitemStop [0]{}%
\providecommand \bibitemNoStop [0]{.\EOS\space}%
\providecommand \EOS [0]{\spacefactor3000\relax}%
\providecommand \BibitemShut  [1]{\csname bibitem#1\endcsname}%
\let\auto@bib@innerbib\@empty
\bibitem [{\citenamefont {Lorenz}(1963)}]{lorenz:1963:deterministic}%
  \BibitemOpen
  \bibfield  {author} {\bibinfo {author} {\bibfnamefont {E.~N.}\ \bibnamefont {Lorenz}},\ }\bibfield  {title} {\bibinfo {title} {Deterministic nonperiodic flow},\ }\href@noop {} {\bibfield  {journal} {\bibinfo  {journal} {Journal of atmospheric sciences}\ }\textbf {\bibinfo {volume} {20}},\ \bibinfo {pages} {130} (\bibinfo {year} {1963})}\BibitemShut {NoStop}%
\bibitem [{\citenamefont {Yang}\ \emph {et~al.}(2007)\citenamefont {Yang}, \citenamefont {Zhang},\ and\ \citenamefont {Li}}]{yang:2007:chaotic}%
  \BibitemOpen
  \bibfield  {author} {\bibinfo {author} {\bibfnamefont {Z.}~\bibnamefont {Yang}}, \bibinfo {author} {\bibfnamefont {S.}~\bibnamefont {Zhang}},\ and\ \bibinfo {author} {\bibfnamefont {Y.~C.}\ \bibnamefont {Li}},\ }\bibfield  {title} {\bibinfo {title} {Chaotic dynamics of spin-valve oscillators},\ }\href@noop {} {\bibfield  {journal} {\bibinfo  {journal} {Physical review letters}\ }\textbf {\bibinfo {volume} {99}},\ \bibinfo {pages} {134101} (\bibinfo {year} {2007})}\BibitemShut {NoStop}%
\bibitem [{\citenamefont {Schiff}\ \emph {et~al.}(1994)\citenamefont {Schiff}, \citenamefont {Jerger}, \citenamefont {Duong}, \citenamefont {Chang}, \citenamefont {Spano},\ and\ \citenamefont {Ditto}}]{schiff:1994:controlling}%
  \BibitemOpen
  \bibfield  {author} {\bibinfo {author} {\bibfnamefont {S.~J.}\ \bibnamefont {Schiff}}, \bibinfo {author} {\bibfnamefont {K.}~\bibnamefont {Jerger}}, \bibinfo {author} {\bibfnamefont {D.~H.}\ \bibnamefont {Duong}}, \bibinfo {author} {\bibfnamefont {T.}~\bibnamefont {Chang}}, \bibinfo {author} {\bibfnamefont {M.~L.}\ \bibnamefont {Spano}},\ and\ \bibinfo {author} {\bibfnamefont {W.~L.}\ \bibnamefont {Ditto}},\ }\bibfield  {title} {\bibinfo {title} {Controlling chaos in the brain},\ }\href@noop {} {\bibfield  {journal} {\bibinfo  {journal} {Nature}\ }\textbf {\bibinfo {volume} {370}},\ \bibinfo {pages} {615} (\bibinfo {year} {1994})}\BibitemShut {NoStop}%
\bibitem [{\citenamefont {Korn}\ and\ \citenamefont {Faure}(2003)}]{korn:2003:there}%
  \BibitemOpen
  \bibfield  {author} {\bibinfo {author} {\bibfnamefont {H.}~\bibnamefont {Korn}}\ and\ \bibinfo {author} {\bibfnamefont {P.}~\bibnamefont {Faure}},\ }\bibfield  {title} {\bibinfo {title} {Is there chaos in the brain? ii. experimental evidence and related models},\ }\href@noop {} {\bibfield  {journal} {\bibinfo  {journal} {Comptes rendus biologies}\ }\textbf {\bibinfo {volume} {326}},\ \bibinfo {pages} {787} (\bibinfo {year} {2003})}\BibitemShut {NoStop}%
\bibitem [{\citenamefont {Pritchard}\ and\ \citenamefont {Duke}(1995)}]{pritchard:1995:measuring}%
  \BibitemOpen
  \bibfield  {author} {\bibinfo {author} {\bibfnamefont {W.~S.}\ \bibnamefont {Pritchard}}\ and\ \bibinfo {author} {\bibfnamefont {D.~W.}\ \bibnamefont {Duke}},\ }\bibfield  {title} {\bibinfo {title} {Measuring chaos in the brain-a tutorial review of eeg dimension estimation},\ }\href@noop {} {\bibfield  {journal} {\bibinfo  {journal} {Brain and cognition}\ }\textbf {\bibinfo {volume} {27}},\ \bibinfo {pages} {353} (\bibinfo {year} {1995})}\BibitemShut {NoStop}%
\bibitem [{\citenamefont {Deco}\ \emph {et~al.}(2011)\citenamefont {Deco}, \citenamefont {Jirsa},\ and\ \citenamefont {McIntosh}}]{deco:2011:emerging}%
  \BibitemOpen
  \bibfield  {author} {\bibinfo {author} {\bibfnamefont {G.}~\bibnamefont {Deco}}, \bibinfo {author} {\bibfnamefont {V.~K.}\ \bibnamefont {Jirsa}},\ and\ \bibinfo {author} {\bibfnamefont {A.~R.}\ \bibnamefont {McIntosh}},\ }\bibfield  {title} {\bibinfo {title} {Emerging concepts for the dynamical organization of resting-state activity in the brain},\ }\href@noop {} {\bibfield  {journal} {\bibinfo  {journal} {Nature Reviews Neuroscience}\ }\textbf {\bibinfo {volume} {12}},\ \bibinfo {pages} {43} (\bibinfo {year} {2011})}\BibitemShut {NoStop}%
\bibitem [{\citenamefont {Backus}(1978)}]{backus:1978:can}%
  \BibitemOpen
  \bibfield  {author} {\bibinfo {author} {\bibfnamefont {J.}~\bibnamefont {Backus}},\ }\bibfield  {title} {\bibinfo {title} {Can programming be liberated from the von neumann style? a functional style and its algebra of programs},\ }\href@noop {} {\bibfield  {journal} {\bibinfo  {journal} {Communications of the ACM}\ }\textbf {\bibinfo {volume} {21}},\ \bibinfo {pages} {613} (\bibinfo {year} {1978})}\BibitemShut {NoStop}%
\bibitem [{\citenamefont {Markovi{\'c}}\ \emph {et~al.}(2020)\citenamefont {Markovi{\'c}}, \citenamefont {Mizrahi}, \citenamefont {Querlioz},\ and\ \citenamefont {Grollier}}]{markovic:2020:physics}%
  \BibitemOpen
  \bibfield  {author} {\bibinfo {author} {\bibfnamefont {D.}~\bibnamefont {Markovi{\'c}}}, \bibinfo {author} {\bibfnamefont {A.}~\bibnamefont {Mizrahi}}, \bibinfo {author} {\bibfnamefont {D.}~\bibnamefont {Querlioz}},\ and\ \bibinfo {author} {\bibfnamefont {J.}~\bibnamefont {Grollier}},\ }\bibfield  {title} {\bibinfo {title} {Physics for neuromorphic computing},\ }\href@noop {} {\bibfield  {journal} {\bibinfo  {journal} {Nature Reviews Physics}\ }\textbf {\bibinfo {volume} {2}},\ \bibinfo {pages} {499} (\bibinfo {year} {2020})}\BibitemShut {NoStop}%
\bibitem [{\citenamefont {Jaeger}\ and\ \citenamefont {Haas}(2004)}]{jaeger:2004:harnessing}%
  \BibitemOpen
  \bibfield  {author} {\bibinfo {author} {\bibfnamefont {H.}~\bibnamefont {Jaeger}}\ and\ \bibinfo {author} {\bibfnamefont {H.}~\bibnamefont {Haas}},\ }\bibfield  {title} {\bibinfo {title} {Harnessing nonlinearity: Predicting chaotic systems and saving energy in wireless communication},\ }\href@noop {} {\bibfield  {journal} {\bibinfo  {journal} {science}\ }\textbf {\bibinfo {volume} {304}},\ \bibinfo {pages} {78} (\bibinfo {year} {2004})}\BibitemShut {NoStop}%
\bibitem [{\citenamefont {Maass}\ \emph {et~al.}(2002)\citenamefont {Maass}, \citenamefont {Natschl{\"a}ger},\ and\ \citenamefont {Markram}}]{maass:2002:real}%
  \BibitemOpen
  \bibfield  {author} {\bibinfo {author} {\bibfnamefont {W.}~\bibnamefont {Maass}}, \bibinfo {author} {\bibfnamefont {T.}~\bibnamefont {Natschl{\"a}ger}},\ and\ \bibinfo {author} {\bibfnamefont {H.}~\bibnamefont {Markram}},\ }\bibfield  {title} {\bibinfo {title} {Real-time computing without stable states: A new framework for neural computation based on perturbations},\ }\href@noop {} {\bibfield  {journal} {\bibinfo  {journal} {Neural computation}\ }\textbf {\bibinfo {volume} {14}},\ \bibinfo {pages} {2531} (\bibinfo {year} {2002})}\BibitemShut {NoStop}%
\bibitem [{\citenamefont {Nakajima}\ and\ \citenamefont {Fischer}(2021)}]{nakajima:2021:reservoir}%
  \BibitemOpen
  \bibfield  {author} {\bibinfo {author} {\bibfnamefont {K.}~\bibnamefont {Nakajima}}\ and\ \bibinfo {author} {\bibfnamefont {I.}~\bibnamefont {Fischer}},\ }\href@noop {} {\emph {\bibinfo {title} {Reservoir computing}}}\ (\bibinfo  {publisher} {Springer},\ \bibinfo {year} {2021})\BibitemShut {NoStop}%
\bibitem [{\citenamefont {Ishihara}\ and\ \citenamefont {Kaneko}(2005)}]{ishihara:2005:magic}%
  \BibitemOpen
  \bibfield  {author} {\bibinfo {author} {\bibfnamefont {S.}~\bibnamefont {Ishihara}}\ and\ \bibinfo {author} {\bibfnamefont {K.}~\bibnamefont {Kaneko}},\ }\bibfield  {title} {\bibinfo {title} {Magic number 7$\pm$2 in networks of threshold dynamics},\ }\href@noop {} {\bibfield  {journal} {\bibinfo  {journal} {Physical review letters}\ }\textbf {\bibinfo {volume} {94}},\ \bibinfo {pages} {058102} (\bibinfo {year} {2005})}\BibitemShut {NoStop}%
\bibitem [{\citenamefont {Saxe}\ \emph {et~al.}(2013)\citenamefont {Saxe}, \citenamefont {McClelland},\ and\ \citenamefont {Ganguli}}]{saxe:2013:exact}%
  \BibitemOpen
  \bibfield  {author} {\bibinfo {author} {\bibfnamefont {A.~M.}\ \bibnamefont {Saxe}}, \bibinfo {author} {\bibfnamefont {J.~L.}\ \bibnamefont {McClelland}},\ and\ \bibinfo {author} {\bibfnamefont {S.}~\bibnamefont {Ganguli}},\ }\bibfield  {title} {\bibinfo {title} {Exact solutions to the nonlinear dynamics of learning in deep linear neural networks},\ }\href@noop {} {\bibfield  {journal} {\bibinfo  {journal} {arXiv preprint arXiv:1312.6120}\ } (\bibinfo {year} {2013})}\BibitemShut {NoStop}%
\bibitem [{\citenamefont {Mototake}\ and\ \citenamefont {Ikegami}(2015)}]{mototake:2015:dynamics}%
  \BibitemOpen
  \bibfield  {author} {\bibinfo {author} {\bibfnamefont {Y.}~\bibnamefont {Mototake}}\ and\ \bibinfo {author} {\bibfnamefont {T.}~\bibnamefont {Ikegami}},\ }\bibfield  {title} {\bibinfo {title} {The dynamics of deep neural networks},\ }in\ \href@noop {} {\emph {\bibinfo {booktitle} {Proceedings of the Twentieth International Symposium on Artificial Life and Robotics}}},\ Vol.~\bibinfo {volume} {20}\ (\bibinfo {year} {2015})\BibitemShut {NoStop}%
\bibitem [{\citenamefont {Schoenholz}\ \emph {et~al.}(2016)\citenamefont {Schoenholz}, \citenamefont {Gilmer}, \citenamefont {Ganguli},\ and\ \citenamefont {Sohl-Dickstein}}]{schoenholz:2016:deep}%
  \BibitemOpen
  \bibfield  {author} {\bibinfo {author} {\bibfnamefont {S.~S.}\ \bibnamefont {Schoenholz}}, \bibinfo {author} {\bibfnamefont {J.}~\bibnamefont {Gilmer}}, \bibinfo {author} {\bibfnamefont {S.}~\bibnamefont {Ganguli}},\ and\ \bibinfo {author} {\bibfnamefont {J.}~\bibnamefont {Sohl-Dickstein}},\ }\bibfield  {title} {\bibinfo {title} {Deep information propagation},\ }\href@noop {} {\bibfield  {journal} {\bibinfo  {journal} {arXiv preprint arXiv:1611.01232}\ } (\bibinfo {year} {2016})}\BibitemShut {NoStop}%
\bibitem [{\citenamefont {Poole}\ \emph {et~al.}(2016)\citenamefont {Poole}, \citenamefont {Lahiri}, \citenamefont {Raghu}, \citenamefont {Sohl-Dickstein},\ and\ \citenamefont {Ganguli}}]{poole:2016:exponential}%
  \BibitemOpen
  \bibfield  {author} {\bibinfo {author} {\bibfnamefont {B.}~\bibnamefont {Poole}}, \bibinfo {author} {\bibfnamefont {S.}~\bibnamefont {Lahiri}}, \bibinfo {author} {\bibfnamefont {M.}~\bibnamefont {Raghu}}, \bibinfo {author} {\bibfnamefont {J.}~\bibnamefont {Sohl-Dickstein}},\ and\ \bibinfo {author} {\bibfnamefont {S.}~\bibnamefont {Ganguli}},\ }\bibfield  {title} {\bibinfo {title} {Exponential expressivity in deep neural networks through transient chaos},\ }\href@noop {} {\bibfield  {journal} {\bibinfo  {journal} {Advances in neural information processing systems}\ }\textbf {\bibinfo {volume} {29}} (\bibinfo {year} {2016})}\BibitemShut {NoStop}%
\bibitem [{\citenamefont {Lin}\ \emph {et~al.}(2020)\citenamefont {Lin}, \citenamefont {Wang}, \citenamefont {Yao},\ and\ \citenamefont {Tan}}]{lin:2020:chaotic}%
  \BibitemOpen
  \bibfield  {author} {\bibinfo {author} {\bibfnamefont {H.}~\bibnamefont {Lin}}, \bibinfo {author} {\bibfnamefont {C.}~\bibnamefont {Wang}}, \bibinfo {author} {\bibfnamefont {W.}~\bibnamefont {Yao}},\ and\ \bibinfo {author} {\bibfnamefont {Y.}~\bibnamefont {Tan}},\ }\bibfield  {title} {\bibinfo {title} {Chaotic dynamics in a neural network with different types of external stimuli},\ }\href@noop {} {\bibfield  {journal} {\bibinfo  {journal} {Communications in Nonlinear Science and Numerical Simulation}\ }\textbf {\bibinfo {volume} {90}},\ \bibinfo {pages} {105390} (\bibinfo {year} {2020})}\BibitemShut {NoStop}%
\bibitem [{\citenamefont {Keup}\ \emph {et~al.}(2021)\citenamefont {Keup}, \citenamefont {K{\"u}hn}, \citenamefont {Dahmen},\ and\ \citenamefont {Helias}}]{keup:2021:transient}%
  \BibitemOpen
  \bibfield  {author} {\bibinfo {author} {\bibfnamefont {C.}~\bibnamefont {Keup}}, \bibinfo {author} {\bibfnamefont {T.}~\bibnamefont {K{\"u}hn}}, \bibinfo {author} {\bibfnamefont {D.}~\bibnamefont {Dahmen}},\ and\ \bibinfo {author} {\bibfnamefont {M.}~\bibnamefont {Helias}},\ }\bibfield  {title} {\bibinfo {title} {Transient chaotic dimensionality expansion by recurrent networks},\ }\href@noop {} {\bibfield  {journal} {\bibinfo  {journal} {Physical Review X}\ }\textbf {\bibinfo {volume} {11}},\ \bibinfo {pages} {021064} (\bibinfo {year} {2021})}\BibitemShut {NoStop}%
\bibitem [{\citenamefont {Inoue}\ \emph {et~al.}(2022)\citenamefont {Inoue}, \citenamefont {Ohara}, \citenamefont {Kuniyoshi},\ and\ \citenamefont {Nakajima}}]{inoue:2022:transient}%
  \BibitemOpen
  \bibfield  {author} {\bibinfo {author} {\bibfnamefont {K.}~\bibnamefont {Inoue}}, \bibinfo {author} {\bibfnamefont {S.}~\bibnamefont {Ohara}}, \bibinfo {author} {\bibfnamefont {Y.}~\bibnamefont {Kuniyoshi}},\ and\ \bibinfo {author} {\bibfnamefont {K.}~\bibnamefont {Nakajima}},\ }\bibfield  {title} {\bibinfo {title} {Transient chaos in bidirectional encoder representations from transformers},\ }\href@noop {} {\bibfield  {journal} {\bibinfo  {journal} {Physical Review Research}\ }\textbf {\bibinfo {volume} {4}},\ \bibinfo {pages} {013204} (\bibinfo {year} {2022})}\BibitemShut {NoStop}%
\bibitem [{\citenamefont {Engelken}\ \emph {et~al.}(2023)\citenamefont {Engelken}, \citenamefont {Wolf},\ and\ \citenamefont {Abbott}}]{engelken:2023:lyapunov}%
  \BibitemOpen
  \bibfield  {author} {\bibinfo {author} {\bibfnamefont {R.}~\bibnamefont {Engelken}}, \bibinfo {author} {\bibfnamefont {F.}~\bibnamefont {Wolf}},\ and\ \bibinfo {author} {\bibfnamefont {L.~F.}\ \bibnamefont {Abbott}},\ }\bibfield  {title} {\bibinfo {title} {Lyapunov spectra of chaotic recurrent neural networks},\ }\href@noop {} {\bibfield  {journal} {\bibinfo  {journal} {Physical Review Research}\ }\textbf {\bibinfo {volume} {5}},\ \bibinfo {pages} {043044} (\bibinfo {year} {2023})}\BibitemShut {NoStop}%
\bibitem [{\citenamefont {Kondo}\ \emph {et~al.}(2021)\citenamefont {Kondo}, \citenamefont {Sunada},\ and\ \citenamefont {Niiyama}}]{Misaki:Kondo:2021}%
  \BibitemOpen
  \bibfield  {author} {\bibinfo {author} {\bibfnamefont {M.}~\bibnamefont {Kondo}}, \bibinfo {author} {\bibfnamefont {S.}~\bibnamefont {Sunada}},\ and\ \bibinfo {author} {\bibfnamefont {T.}~\bibnamefont {Niiyama}},\ }\bibfield  {title} {\bibinfo {title} {Lyapunov exponent analysis for multilayer neural networks},\ }\href {https://doi.org/10.1587/nolta.12.674} {\bibfield  {journal} {\bibinfo  {journal} {Nonlinear Theory and Its Applications, IEICE}\ }\textbf {\bibinfo {volume} {12}},\ \bibinfo {pages} {674} (\bibinfo {year} {2021})}\BibitemShut {NoStop}%
\bibitem [{\citenamefont {He}\ \emph {et~al.}(2016)\citenamefont {He}, \citenamefont {Zhang}, \citenamefont {Ren},\ and\ \citenamefont {Sun}}]{he:2016:deep}%
  \BibitemOpen
  \bibfield  {author} {\bibinfo {author} {\bibfnamefont {K.}~\bibnamefont {He}}, \bibinfo {author} {\bibfnamefont {X.}~\bibnamefont {Zhang}}, \bibinfo {author} {\bibfnamefont {S.}~\bibnamefont {Ren}},\ and\ \bibinfo {author} {\bibfnamefont {J.}~\bibnamefont {Sun}},\ }\bibfield  {title} {\bibinfo {title} {Deep residual learning for image recognition},\ }in\ \href@noop {} {\emph {\bibinfo {booktitle} {Proceedings of the IEEE conference on computer vision and pattern recognition}}}\ (\bibinfo {year} {2016})\ pp.\ \bibinfo {pages} {770--778}\BibitemShut {NoStop}%
\bibitem [{\citenamefont {Deng}\ \emph {et~al.}(2009)\citenamefont {Deng}, \citenamefont {Dong}, \citenamefont {Socher}, \citenamefont {Li}, \citenamefont {Li},\ and\ \citenamefont {Fei-Fei}}]{deng:imagenet:2009}%
  \BibitemOpen
  \bibfield  {author} {\bibinfo {author} {\bibfnamefont {J.}~\bibnamefont {Deng}}, \bibinfo {author} {\bibfnamefont {W.}~\bibnamefont {Dong}}, \bibinfo {author} {\bibfnamefont {R.}~\bibnamefont {Socher}}, \bibinfo {author} {\bibfnamefont {L.-J.}\ \bibnamefont {Li}}, \bibinfo {author} {\bibfnamefont {K.}~\bibnamefont {Li}},\ and\ \bibinfo {author} {\bibfnamefont {L.}~\bibnamefont {Fei-Fei}},\ }\bibfield  {title} {\bibinfo {title} {Imagenet: A large-scale hierarchical image database},\ }in\ \href {https://doi.org/10.1109/CVPR.2009.5206848} {\emph {\bibinfo {booktitle} {2009 IEEE Conference on Computer Vision and Pattern Recognition}}}\ (\bibinfo {year} {2009})\ pp.\ \bibinfo {pages} {248--255}\BibitemShut {NoStop}%
\bibitem [{\citenamefont {Krizhevsky}\ \emph {et~al.}(2009)\citenamefont {Krizhevsky}, \citenamefont {Hinton} \emph {et~al.}}]{krizhevsky:2009:learning}%
  \BibitemOpen
  \bibfield  {author} {\bibinfo {author} {\bibfnamefont {A.}~\bibnamefont {Krizhevsky}}, \bibinfo {author} {\bibfnamefont {G.}~\bibnamefont {Hinton}}, \emph {et~al.},\ }\href@noop {} {\emph {\bibinfo {title} {Learning multiple layers of features from tiny images}}},\ \bibinfo {type} {Tech. Rep.}\ (\bibinfo  {institution} {University of Toronto},\ \bibinfo {address} {Toronto, ON, Canada},\ \bibinfo {year} {2009})\BibitemShut {NoStop}%
\bibitem [{\citenamefont {Devlin}\ \emph {et~al.}(2018)\citenamefont {Devlin}, \citenamefont {Chang}, \citenamefont {Lee},\ and\ \citenamefont {Toutanova}}]{devlin:2018:bert}%
  \BibitemOpen
  \bibfield  {author} {\bibinfo {author} {\bibfnamefont {J.}~\bibnamefont {Devlin}}, \bibinfo {author} {\bibfnamefont {M.-W.}\ \bibnamefont {Chang}}, \bibinfo {author} {\bibfnamefont {K.}~\bibnamefont {Lee}},\ and\ \bibinfo {author} {\bibfnamefont {K.}~\bibnamefont {Toutanova}},\ }\bibfield  {title} {\bibinfo {title} {Bert: Pre-training of deep bidirectional transformers for language understanding},\ }\href@noop {} {\bibfield  {journal} {\bibinfo  {journal} {arXiv preprint arXiv:1810.04805}\ } (\bibinfo {year} {2018})}\BibitemShut {NoStop}%
\bibitem [{\citenamefont {Turc}\ \emph {et~al.}(2019)\citenamefont {Turc}, \citenamefont {Chang}, \citenamefont {Lee},\ and\ \citenamefont {Toutanova}}]{turc:2019:well}%
  \BibitemOpen
  \bibfield  {author} {\bibinfo {author} {\bibfnamefont {I.}~\bibnamefont {Turc}}, \bibinfo {author} {\bibfnamefont {M.-W.}\ \bibnamefont {Chang}}, \bibinfo {author} {\bibfnamefont {K.}~\bibnamefont {Lee}},\ and\ \bibinfo {author} {\bibfnamefont {K.}~\bibnamefont {Toutanova}},\ }\bibfield  {title} {\bibinfo {title} {Well-read students learn better: On the importance of pre-training compact models},\ }\href@noop {} {\bibfield  {journal} {\bibinfo  {journal} {arXiv preprint arXiv:1908.08962v2}\ } (\bibinfo {year} {2019})}\BibitemShut {NoStop}%
\bibitem [{\citenamefont {Maas}\ \emph {et~al.}(2011)\citenamefont {Maas}, \citenamefont {Daly}, \citenamefont {Pham}, \citenamefont {Huang}, \citenamefont {Ng},\ and\ \citenamefont {Potts}}]{maas:2011:learning}%
  \BibitemOpen
  \bibfield  {author} {\bibinfo {author} {\bibfnamefont {A.~L.}\ \bibnamefont {Maas}}, \bibinfo {author} {\bibfnamefont {R.~E.}\ \bibnamefont {Daly}}, \bibinfo {author} {\bibfnamefont {P.~T.}\ \bibnamefont {Pham}}, \bibinfo {author} {\bibfnamefont {D.}~\bibnamefont {Huang}}, \bibinfo {author} {\bibfnamefont {A.~Y.}\ \bibnamefont {Ng}},\ and\ \bibinfo {author} {\bibfnamefont {C.}~\bibnamefont {Potts}},\ }\bibfield  {title} {\bibinfo {title} {Learning word vectors for sentiment analysis},\ }in\ \href {https://aclanthology.org/P11-1015} {\emph {\bibinfo {booktitle} {Proceedings of the 49th Annual Meeting of the Association for Computational Linguistics: Human Language Technologies}}}\ (\bibinfo  {publisher} {Association for Computational Linguistics},\ \bibinfo {address} {Portland, Oregon, USA},\ \bibinfo {year} {2011})\ pp.\ \bibinfo {pages} {142--150}\BibitemShut {NoStop}%
\bibitem [{\citenamefont {Sompolinsky}\ \emph {et~al.}(1988)\citenamefont {Sompolinsky}, \citenamefont {Crisanti},\ and\ \citenamefont {Sommers}}]{sompolinsky:1988:chaos}%
  \BibitemOpen
  \bibfield  {author} {\bibinfo {author} {\bibfnamefont {H.}~\bibnamefont {Sompolinsky}}, \bibinfo {author} {\bibfnamefont {A.}~\bibnamefont {Crisanti}},\ and\ \bibinfo {author} {\bibfnamefont {H.-J.}\ \bibnamefont {Sommers}},\ }\bibfield  {title} {\bibinfo {title} {Chaos in random neural networks},\ }\href@noop {} {\bibfield  {journal} {\bibinfo  {journal} {Physical review letters}\ }\textbf {\bibinfo {volume} {61}},\ \bibinfo {pages} {259} (\bibinfo {year} {1988})}\BibitemShut {NoStop}%
\bibitem [{\citenamefont {Verstraeten}\ \emph {et~al.}(2007)\citenamefont {Verstraeten}, \citenamefont {Schrauwen}, \citenamefont {d’Haene},\ and\ \citenamefont {Stroobandt}}]{verstraeten2007experimental}%
  \BibitemOpen
  \bibfield  {author} {\bibinfo {author} {\bibfnamefont {D.}~\bibnamefont {Verstraeten}}, \bibinfo {author} {\bibfnamefont {B.}~\bibnamefont {Schrauwen}}, \bibinfo {author} {\bibfnamefont {M.}~\bibnamefont {d’Haene}},\ and\ \bibinfo {author} {\bibfnamefont {D.}~\bibnamefont {Stroobandt}},\ }\bibfield  {title} {\bibinfo {title} {An experimental unification of reservoir computing methods},\ }\href@noop {} {\bibfield  {journal} {\bibinfo  {journal} {Neural networks}\ }\textbf {\bibinfo {volume} {20}},\ \bibinfo {pages} {391} (\bibinfo {year} {2007})}\BibitemShut {NoStop}%
\bibitem [{\citenamefont {Lorenz}(1996)}]{lorenz:1996:predictability}%
  \BibitemOpen
  \bibfield  {author} {\bibinfo {author} {\bibfnamefont {E.~N.}\ \bibnamefont {Lorenz}},\ }\bibfield  {title} {\bibinfo {title} {Predictability: A problem partly solved},\ }in\ \href@noop {} {\emph {\bibinfo {booktitle} {Proc. Seminar on predictability}}},\ Vol.~\bibinfo {volume} {1}\ (\bibinfo {organization} {Reading},\ \bibinfo {year} {1996})\BibitemShut {NoStop}%
\bibitem [{\citenamefont {Pontryagin}(1987)}]{pontryagin:1987:mathematical}%
  \BibitemOpen
  \bibfield  {author} {\bibinfo {author} {\bibfnamefont {L.~S.}\ \bibnamefont {Pontryagin}},\ }\href@noop {} {\emph {\bibinfo {title} {Mathematical theory of optimal processes}}}\ (\bibinfo  {publisher} {CRC press},\ \bibinfo {year} {1987})\BibitemShut {NoStop}%
\bibitem [{\citenamefont {Chen}\ \emph {et~al.}(2018)\citenamefont {Chen}, \citenamefont {Rubanova}, \citenamefont {Bettencourt},\ and\ \citenamefont {Duvenaud}}]{chen:2018:advances}%
  \BibitemOpen
  \bibfield  {author} {\bibinfo {author} {\bibfnamefont {R.~T.~Q.}\ \bibnamefont {Chen}}, \bibinfo {author} {\bibfnamefont {Y.}~\bibnamefont {Rubanova}}, \bibinfo {author} {\bibfnamefont {J.}~\bibnamefont {Bettencourt}},\ and\ \bibinfo {author} {\bibfnamefont {D.~K.}\ \bibnamefont {Duvenaud}},\ }\bibfield  {title} {\bibinfo {title} {Neural ordinary differential equations},\ }in\ \href {https://proceedings.neurips.cc/paper_files/paper/2018/file/69386f6bb1dfed68692a24c8686939b9-Paper.pdf} {\emph {\bibinfo {booktitle} {Advances in Neural Information Processing Systems}}},\ Vol.~\bibinfo {volume} {31},\ \bibinfo {editor} {edited by\ \bibinfo {editor} {\bibfnamefont {S.}~\bibnamefont {Bengio}}, \bibinfo {editor} {\bibfnamefont {H.}~\bibnamefont {Wallach}}, \bibinfo {editor} {\bibfnamefont {H.}~\bibnamefont {Larochelle}}, \bibinfo {editor} {\bibfnamefont {K.}~\bibnamefont {Grauman}}, \bibinfo {editor} {\bibfnamefont {N.}~\bibnamefont {Cesa-Bianchi}},\ and\ \bibinfo {editor} {\bibfnamefont {R.}~\bibnamefont
  {Garnett}}}\ (\bibinfo  {publisher} {Curran Associates, Inc.},\ \bibinfo {year} {2018})\BibitemShut {NoStop}%
\bibitem [{\citenamefont {Torrejon}\ \emph {et~al.}(2017)\citenamefont {Torrejon}, \citenamefont {Riou}, \citenamefont {Araujo}, \citenamefont {Tsunegi}, \citenamefont {Khalsa}, \citenamefont {Querlioz}, \citenamefont {Bortolotti}, \citenamefont {Cros}, \citenamefont {Yakushiji}, \citenamefont {Fukushima} \emph {et~al.}}]{torrejon:2017:neuromorphic}%
  \BibitemOpen
  \bibfield  {author} {\bibinfo {author} {\bibfnamefont {J.}~\bibnamefont {Torrejon}}, \bibinfo {author} {\bibfnamefont {M.}~\bibnamefont {Riou}}, \bibinfo {author} {\bibfnamefont {F.~A.}\ \bibnamefont {Araujo}}, \bibinfo {author} {\bibfnamefont {S.}~\bibnamefont {Tsunegi}}, \bibinfo {author} {\bibfnamefont {G.}~\bibnamefont {Khalsa}}, \bibinfo {author} {\bibfnamefont {D.}~\bibnamefont {Querlioz}}, \bibinfo {author} {\bibfnamefont {P.}~\bibnamefont {Bortolotti}}, \bibinfo {author} {\bibfnamefont {V.}~\bibnamefont {Cros}}, \bibinfo {author} {\bibfnamefont {K.}~\bibnamefont {Yakushiji}}, \bibinfo {author} {\bibfnamefont {A.}~\bibnamefont {Fukushima}}, \emph {et~al.},\ }\bibfield  {title} {\bibinfo {title} {Neuromorphic computing with nanoscale spintronic oscillators},\ }\href@noop {} {\bibfield  {journal} {\bibinfo  {journal} {Nature}\ }\textbf {\bibinfo {volume} {547}},\ \bibinfo {pages} {428} (\bibinfo {year} {2017})}\BibitemShut {NoStop}%
\bibitem [{\citenamefont {Tsunegi}\ \emph {et~al.}(2019)\citenamefont {Tsunegi}, \citenamefont {Taniguchi}, \citenamefont {Nakajima}, \citenamefont {Miwa}, \citenamefont {Yakushiji}, \citenamefont {Fukushima}, \citenamefont {Yuasa},\ and\ \citenamefont {Kubota}}]{tsunegi:2019:physical}%
  \BibitemOpen
  \bibfield  {author} {\bibinfo {author} {\bibfnamefont {S.}~\bibnamefont {Tsunegi}}, \bibinfo {author} {\bibfnamefont {T.}~\bibnamefont {Taniguchi}}, \bibinfo {author} {\bibfnamefont {K.}~\bibnamefont {Nakajima}}, \bibinfo {author} {\bibfnamefont {S.}~\bibnamefont {Miwa}}, \bibinfo {author} {\bibfnamefont {K.}~\bibnamefont {Yakushiji}}, \bibinfo {author} {\bibfnamefont {A.}~\bibnamefont {Fukushima}}, \bibinfo {author} {\bibfnamefont {S.}~\bibnamefont {Yuasa}},\ and\ \bibinfo {author} {\bibfnamefont {H.}~\bibnamefont {Kubota}},\ }\bibfield  {title} {\bibinfo {title} {Physical reservoir computing based on spin torque oscillator with forced synchronization},\ }\href@noop {} {\bibfield  {journal} {\bibinfo  {journal} {Applied Physics Letters}\ }\textbf {\bibinfo {volume} {114}} (\bibinfo {year} {2019})}\BibitemShut {NoStop}%
\bibitem [{\citenamefont {Grollier}\ \emph {et~al.}(2020)\citenamefont {Grollier}, \citenamefont {Querlioz}, \citenamefont {Camsari}, \citenamefont {Everschor-Sitte}, \citenamefont {Fukami},\ and\ \citenamefont {Stiles}}]{grollier:2020:neuromorphic}%
  \BibitemOpen
  \bibfield  {author} {\bibinfo {author} {\bibfnamefont {J.}~\bibnamefont {Grollier}}, \bibinfo {author} {\bibfnamefont {D.}~\bibnamefont {Querlioz}}, \bibinfo {author} {\bibfnamefont {K.}~\bibnamefont {Camsari}}, \bibinfo {author} {\bibfnamefont {K.}~\bibnamefont {Everschor-Sitte}}, \bibinfo {author} {\bibfnamefont {S.}~\bibnamefont {Fukami}},\ and\ \bibinfo {author} {\bibfnamefont {M.~D.}\ \bibnamefont {Stiles}},\ }\bibfield  {title} {\bibinfo {title} {Neuromorphic spintronics},\ }\href@noop {} {\bibfield  {journal} {\bibinfo  {journal} {Nature electronics}\ }\textbf {\bibinfo {volume} {3}},\ \bibinfo {pages} {360} (\bibinfo {year} {2020})}\BibitemShut {NoStop}%
\bibitem [{\citenamefont {Yamaguchi}\ \emph {et~al.}(2019)\citenamefont {Yamaguchi}, \citenamefont {Akashi}, \citenamefont {Nakajima}, \citenamefont {Tsunegi}, \citenamefont {Kubota},\ and\ \citenamefont {Taniguchi}}]{yamaguchi:2019:synchronization}%
  \BibitemOpen
  \bibfield  {author} {\bibinfo {author} {\bibfnamefont {T.}~\bibnamefont {Yamaguchi}}, \bibinfo {author} {\bibfnamefont {N.}~\bibnamefont {Akashi}}, \bibinfo {author} {\bibfnamefont {K.}~\bibnamefont {Nakajima}}, \bibinfo {author} {\bibfnamefont {S.}~\bibnamefont {Tsunegi}}, \bibinfo {author} {\bibfnamefont {H.}~\bibnamefont {Kubota}},\ and\ \bibinfo {author} {\bibfnamefont {T.}~\bibnamefont {Taniguchi}},\ }\bibfield  {title} {\bibinfo {title} {Synchronization and chaos in a spin-torque oscillator with a perpendicularly magnetized free layer},\ }\href@noop {} {\bibfield  {journal} {\bibinfo  {journal} {Physical Review B}\ }\textbf {\bibinfo {volume} {100}},\ \bibinfo {pages} {224422} (\bibinfo {year} {2019})}\BibitemShut {NoStop}%
\bibitem [{\citenamefont {Akashi}\ \emph {et~al.}(2020)\citenamefont {Akashi}, \citenamefont {Yamaguchi}, \citenamefont {Tsunegi}, \citenamefont {Taniguchi}, \citenamefont {Nishida}, \citenamefont {Sakurai}, \citenamefont {Wakao},\ and\ \citenamefont {Nakajima}}]{akashi:2020:input}%
  \BibitemOpen
  \bibfield  {author} {\bibinfo {author} {\bibfnamefont {N.}~\bibnamefont {Akashi}}, \bibinfo {author} {\bibfnamefont {T.}~\bibnamefont {Yamaguchi}}, \bibinfo {author} {\bibfnamefont {S.}~\bibnamefont {Tsunegi}}, \bibinfo {author} {\bibfnamefont {T.}~\bibnamefont {Taniguchi}}, \bibinfo {author} {\bibfnamefont {M.}~\bibnamefont {Nishida}}, \bibinfo {author} {\bibfnamefont {R.}~\bibnamefont {Sakurai}}, \bibinfo {author} {\bibfnamefont {Y.}~\bibnamefont {Wakao}},\ and\ \bibinfo {author} {\bibfnamefont {K.}~\bibnamefont {Nakajima}},\ }\bibfield  {title} {\bibinfo {title} {Input-driven bifurcations and information processing capacity in spintronics reservoirs},\ }\href@noop {} {\bibfield  {journal} {\bibinfo  {journal} {Physical Review Research}\ }\textbf {\bibinfo {volume} {2}},\ \bibinfo {pages} {043303} (\bibinfo {year} {2020})}\BibitemShut {NoStop}%
\bibitem [{\citenamefont {Taniguchi}\ \emph {et~al.}(2019)\citenamefont {Taniguchi}, \citenamefont {Akashi}, \citenamefont {Notsu}, \citenamefont {Kimura}, \citenamefont {Tsukahara},\ and\ \citenamefont {Nakajima}}]{taniguchi:2019:chaos}%
  \BibitemOpen
  \bibfield  {author} {\bibinfo {author} {\bibfnamefont {T.}~\bibnamefont {Taniguchi}}, \bibinfo {author} {\bibfnamefont {N.}~\bibnamefont {Akashi}}, \bibinfo {author} {\bibfnamefont {H.}~\bibnamefont {Notsu}}, \bibinfo {author} {\bibfnamefont {M.}~\bibnamefont {Kimura}}, \bibinfo {author} {\bibfnamefont {H.}~\bibnamefont {Tsukahara}},\ and\ \bibinfo {author} {\bibfnamefont {K.}~\bibnamefont {Nakajima}},\ }\bibfield  {title} {\bibinfo {title} {Chaos in nanomagnet via feedback current},\ }\href@noop {} {\bibfield  {journal} {\bibinfo  {journal} {Physical Review B}\ }\textbf {\bibinfo {volume} {100}},\ \bibinfo {pages} {174425} (\bibinfo {year} {2019})}\BibitemShut {NoStop}%
\bibitem [{\citenamefont {Kamimaki}\ \emph {et~al.}(2021)\citenamefont {Kamimaki}, \citenamefont {Kubota}, \citenamefont {Tsunegi}, \citenamefont {Nakajima}, \citenamefont {Taniguchi}, \citenamefont {Grollier}, \citenamefont {Cros}, \citenamefont {Yakushiji}, \citenamefont {Fukushima}, \citenamefont {Yuasa} \emph {et~al.}}]{kamimaki:2021:chaos}%
  \BibitemOpen
  \bibfield  {author} {\bibinfo {author} {\bibfnamefont {A.}~\bibnamefont {Kamimaki}}, \bibinfo {author} {\bibfnamefont {T.}~\bibnamefont {Kubota}}, \bibinfo {author} {\bibfnamefont {S.}~\bibnamefont {Tsunegi}}, \bibinfo {author} {\bibfnamefont {K.}~\bibnamefont {Nakajima}}, \bibinfo {author} {\bibfnamefont {T.}~\bibnamefont {Taniguchi}}, \bibinfo {author} {\bibfnamefont {J.}~\bibnamefont {Grollier}}, \bibinfo {author} {\bibfnamefont {V.}~\bibnamefont {Cros}}, \bibinfo {author} {\bibfnamefont {K.}~\bibnamefont {Yakushiji}}, \bibinfo {author} {\bibfnamefont {A.}~\bibnamefont {Fukushima}}, \bibinfo {author} {\bibfnamefont {S.}~\bibnamefont {Yuasa}}, \emph {et~al.},\ }\bibfield  {title} {\bibinfo {title} {Chaos in spin-torque oscillator with feedback circuit},\ }\href@noop {} {\bibfield  {journal} {\bibinfo  {journal} {Physical Review Research}\ }\textbf {\bibinfo {volume} {3}},\ \bibinfo {pages} {043216} (\bibinfo {year} {2021})}\BibitemShut {NoStop}%
\bibitem [{\citenamefont {Akashi}\ \emph {et~al.}(2022)\citenamefont {Akashi}, \citenamefont {Kuniyoshi}, \citenamefont {Tsunegi}, \citenamefont {Taniguchi}, \citenamefont {Nishida}, \citenamefont {Sakurai}, \citenamefont {Wakao}, \citenamefont {Kawashima},\ and\ \citenamefont {Nakajima}}]{akashi:2022:coupled}%
  \BibitemOpen
  \bibfield  {author} {\bibinfo {author} {\bibfnamefont {N.}~\bibnamefont {Akashi}}, \bibinfo {author} {\bibfnamefont {Y.}~\bibnamefont {Kuniyoshi}}, \bibinfo {author} {\bibfnamefont {S.}~\bibnamefont {Tsunegi}}, \bibinfo {author} {\bibfnamefont {T.}~\bibnamefont {Taniguchi}}, \bibinfo {author} {\bibfnamefont {M.}~\bibnamefont {Nishida}}, \bibinfo {author} {\bibfnamefont {R.}~\bibnamefont {Sakurai}}, \bibinfo {author} {\bibfnamefont {Y.}~\bibnamefont {Wakao}}, \bibinfo {author} {\bibfnamefont {K.}~\bibnamefont {Kawashima}},\ and\ \bibinfo {author} {\bibfnamefont {K.}~\bibnamefont {Nakajima}},\ }\bibfield  {title} {\bibinfo {title} {A coupled spintronics neuromorphic approach for high-performance reservoir computing},\ }\href {https://doi.org/https://doi.org/10.1002/aisy.202200123} {\bibfield  {journal} {\bibinfo  {journal} {Advanced Intelligent Systems}\ }\textbf {\bibinfo {volume} {4}},\ \bibinfo {pages} {2200123} (\bibinfo {year} {2022})},\ \Eprint
  {https://arxiv.org/abs/https://onlinelibrary.wiley.com/doi/pdf/10.1002/aisy.202200123} {https://onlinelibrary.wiley.com/doi/pdf/10.1002/aisy.202200123} \BibitemShut {NoStop}%
\bibitem [{\citenamefont {Leroux}\ \emph {et~al.}(2022)\citenamefont {Leroux}, \citenamefont {Riz}, \citenamefont {Sanz-Hernández}, \citenamefont {Marković}, \citenamefont {Mizrahi},\ and\ \citenamefont {Grollier}}]{leroux:2022:convolutional}%
  \BibitemOpen
  \bibfield  {author} {\bibinfo {author} {\bibfnamefont {N.}~\bibnamefont {Leroux}}, \bibinfo {author} {\bibfnamefont {A.~D.}\ \bibnamefont {Riz}}, \bibinfo {author} {\bibfnamefont {D.}~\bibnamefont {Sanz-Hernández}}, \bibinfo {author} {\bibfnamefont {D.}~\bibnamefont {Marković}}, \bibinfo {author} {\bibfnamefont {A.}~\bibnamefont {Mizrahi}},\ and\ \bibinfo {author} {\bibfnamefont {J.}~\bibnamefont {Grollier}},\ }\bibfield  {title} {\bibinfo {title} {Convolutional neural networks with radio-frequency spintronic nano-devices},\ }\href {https://doi.org/10.1088/2634-4386/ac77b2} {\bibfield  {journal} {\bibinfo  {journal} {Neuromorphic Computing and Engineering}\ }\textbf {\bibinfo {volume} {2}},\ \bibinfo {pages} {034002} (\bibinfo {year} {2022})}\BibitemShut {NoStop}%
\bibitem [{\citenamefont {Kumar}\ \emph {et~al.}(2017)\citenamefont {Kumar}, \citenamefont {Strachan},\ and\ \citenamefont {Williams}}]{kumar:2017:chaotic}%
  \BibitemOpen
  \bibfield  {author} {\bibinfo {author} {\bibfnamefont {S.}~\bibnamefont {Kumar}}, \bibinfo {author} {\bibfnamefont {J.~P.}\ \bibnamefont {Strachan}},\ and\ \bibinfo {author} {\bibfnamefont {R.~S.}\ \bibnamefont {Williams}},\ }\bibfield  {title} {\bibinfo {title} {Chaotic dynamics in nanoscale nbo2 mott memristors for analogue computing},\ }\href@noop {} {\bibfield  {journal} {\bibinfo  {journal} {Nature}\ }\textbf {\bibinfo {volume} {548}},\ \bibinfo {pages} {318} (\bibinfo {year} {2017})}\BibitemShut {NoStop}%
\bibitem [{\citenamefont {Kumar}\ \emph {et~al.}(2020)\citenamefont {Kumar}, \citenamefont {Williams},\ and\ \citenamefont {Wang}}]{kumar:2020:third}%
  \BibitemOpen
  \bibfield  {author} {\bibinfo {author} {\bibfnamefont {S.}~\bibnamefont {Kumar}}, \bibinfo {author} {\bibfnamefont {R.~S.}\ \bibnamefont {Williams}},\ and\ \bibinfo {author} {\bibfnamefont {Z.}~\bibnamefont {Wang}},\ }\bibfield  {title} {\bibinfo {title} {Third-order nanocircuit elements for neuromorphic engineering},\ }\href@noop {} {\bibfield  {journal} {\bibinfo  {journal} {Nature}\ }\textbf {\bibinfo {volume} {585}},\ \bibinfo {pages} {518} (\bibinfo {year} {2020})}\BibitemShut {NoStop}%
\bibitem [{\citenamefont {Molgedey}\ \emph {et~al.}(1992)\citenamefont {Molgedey}, \citenamefont {Schuchhardt},\ and\ \citenamefont {Schuster}}]{molgedey:1992:suppressing}%
  \BibitemOpen
  \bibfield  {author} {\bibinfo {author} {\bibfnamefont {L.}~\bibnamefont {Molgedey}}, \bibinfo {author} {\bibfnamefont {J.}~\bibnamefont {Schuchhardt}},\ and\ \bibinfo {author} {\bibfnamefont {H.~G.}\ \bibnamefont {Schuster}},\ }\bibfield  {title} {\bibinfo {title} {Suppressing chaos in neural networks by noise},\ }\href@noop {} {\bibfield  {journal} {\bibinfo  {journal} {Physical review letters}\ }\textbf {\bibinfo {volume} {69}},\ \bibinfo {pages} {3717} (\bibinfo {year} {1992})}\BibitemShut {NoStop}%
\bibitem [{\citenamefont {Wang}(1996)}]{wang:1996:suppressing}%
  \BibitemOpen
  \bibfield  {author} {\bibinfo {author} {\bibfnamefont {L.}~\bibnamefont {Wang}},\ }\bibfield  {title} {\bibinfo {title} {Suppressing chaos with hysteresis in a higher order neural network},\ }\href@noop {} {\bibfield  {journal} {\bibinfo  {journal} {IEEE Transactions on Circuits and Systems II: Analog and Digital Signal Processing}\ }\textbf {\bibinfo {volume} {43}},\ \bibinfo {pages} {845} (\bibinfo {year} {1996})}\BibitemShut {NoStop}%
\bibitem [{\citenamefont {Rajan}\ \emph {et~al.}(2010)\citenamefont {Rajan}, \citenamefont {Abbott},\ and\ \citenamefont {Sompolinsky}}]{rajan:2010:stimulus}%
  \BibitemOpen
  \bibfield  {author} {\bibinfo {author} {\bibfnamefont {K.}~\bibnamefont {Rajan}}, \bibinfo {author} {\bibfnamefont {L.}~\bibnamefont {Abbott}},\ and\ \bibinfo {author} {\bibfnamefont {H.}~\bibnamefont {Sompolinsky}},\ }\bibfield  {title} {\bibinfo {title} {Stimulus-dependent suppression of chaos in recurrent neural networks},\ }\href@noop {} {\bibfield  {journal} {\bibinfo  {journal} {Physical review e}\ }\textbf {\bibinfo {volume} {82}},\ \bibinfo {pages} {011903} (\bibinfo {year} {2010})}\BibitemShut {NoStop}%
\bibitem [{\citenamefont {Li}\ \emph {et~al.}(2013)\citenamefont {Li}, \citenamefont {Zhu}, \citenamefont {Xie}, \citenamefont {Chen}, \citenamefont {Aihara},\ and\ \citenamefont {He}}]{li:2013:controlling}%
  \BibitemOpen
  \bibfield  {author} {\bibinfo {author} {\bibfnamefont {Y.}~\bibnamefont {Li}}, \bibinfo {author} {\bibfnamefont {P.}~\bibnamefont {Zhu}}, \bibinfo {author} {\bibfnamefont {X.}~\bibnamefont {Xie}}, \bibinfo {author} {\bibfnamefont {H.}~\bibnamefont {Chen}}, \bibinfo {author} {\bibfnamefont {K.}~\bibnamefont {Aihara}},\ and\ \bibinfo {author} {\bibfnamefont {G.}~\bibnamefont {He}},\ }\bibfield  {title} {\bibinfo {title} {Controlling a chaotic neural network for information processing},\ }\href@noop {} {\bibfield  {journal} {\bibinfo  {journal} {Neurocomputing}\ }\textbf {\bibinfo {volume} {110}},\ \bibinfo {pages} {111} (\bibinfo {year} {2013})}\BibitemShut {NoStop}%
\bibitem [{\citenamefont {Allehiany}\ \emph {et~al.}(2021)\citenamefont {Allehiany}, \citenamefont {Mahmoud}, \citenamefont {Jahanzaib}, \citenamefont {Trikha},\ and\ \citenamefont {Alotaibi}}]{allehiany:2021:chaos}%
  \BibitemOpen
  \bibfield  {author} {\bibinfo {author} {\bibfnamefont {F.}~\bibnamefont {Allehiany}}, \bibinfo {author} {\bibfnamefont {E.~E.}\ \bibnamefont {Mahmoud}}, \bibinfo {author} {\bibfnamefont {L.~S.}\ \bibnamefont {Jahanzaib}}, \bibinfo {author} {\bibfnamefont {P.}~\bibnamefont {Trikha}},\ and\ \bibinfo {author} {\bibfnamefont {H.}~\bibnamefont {Alotaibi}},\ }\bibfield  {title} {\bibinfo {title} {Chaos control and analysis of fractional order neural network under electromagnetic radiation},\ }\href@noop {} {\bibfield  {journal} {\bibinfo  {journal} {Results in Physics}\ }\textbf {\bibinfo {volume} {21}},\ \bibinfo {pages} {103786} (\bibinfo {year} {2021})}\BibitemShut {NoStop}%
\bibitem [{\citenamefont {Shaw}(1981)}]{Shaw:1981strange}%
  \BibitemOpen
  \bibfield  {author} {\bibinfo {author} {\bibfnamefont {R.}~\bibnamefont {Shaw}},\ }\bibfield  {title} {\bibinfo {title} {Strange attractors, chaotic behavior, and information flow},\ }\href {https://doi.org/doi:10.1515/zna-1981-0115} {\bibfield  {journal} {\bibinfo  {journal} {Zeitschrift für Naturforschung A}\ }\textbf {\bibinfo {volume} {36}},\ \bibinfo {pages} {80} (\bibinfo {year} {1981})}\BibitemShut {NoStop}%
\bibitem [{\citenamefont {Laurent}\ and\ \citenamefont {von Brecht}(2017)}]{laurent:2017:a}%
  \BibitemOpen
  \bibfield  {author} {\bibinfo {author} {\bibfnamefont {T.}~\bibnamefont {Laurent}}\ and\ \bibinfo {author} {\bibfnamefont {J.}~\bibnamefont {von Brecht}},\ }\bibfield  {title} {\bibinfo {title} {A recurrent neural network without chaos},\ }in\ \href {https://openreview.net/forum?id=S1dIzvclg} {\emph {\bibinfo {booktitle} {International Conference on Learning Representations}}}\ (\bibinfo {year} {2017})\BibitemShut {NoStop}%
\bibitem [{\citenamefont {Chang}\ \emph {et~al.}(2019)\citenamefont {Chang}, \citenamefont {Chen}, \citenamefont {Haber},\ and\ \citenamefont {Chi}}]{chang:2018:antisymmetricrnn}%
  \BibitemOpen
  \bibfield  {author} {\bibinfo {author} {\bibfnamefont {B.}~\bibnamefont {Chang}}, \bibinfo {author} {\bibfnamefont {M.}~\bibnamefont {Chen}}, \bibinfo {author} {\bibfnamefont {E.}~\bibnamefont {Haber}},\ and\ \bibinfo {author} {\bibfnamefont {E.~H.}\ \bibnamefont {Chi}},\ }\bibfield  {title} {\bibinfo {title} {Antisymmetric{RNN}: A dynamical system view on recurrent neural networks},\ }in\ \href {https://openreview.net/forum?id=ryxepo0cFX} {\emph {\bibinfo {booktitle} {International Conference on Learning Representations}}}\ (\bibinfo {year} {2019})\BibitemShut {NoStop}%
\bibitem [{\citenamefont {Erichson}\ \emph {et~al.}(2021{\natexlab{a}})\citenamefont {Erichson}, \citenamefont {Azencot}, \citenamefont {Queiruga}, \citenamefont {Hodgkinson},\ and\ \citenamefont {Mahoney}}]{erichson:2021:lipschitz}%
  \BibitemOpen
  \bibfield  {author} {\bibinfo {author} {\bibfnamefont {N.~B.}\ \bibnamefont {Erichson}}, \bibinfo {author} {\bibfnamefont {O.}~\bibnamefont {Azencot}}, \bibinfo {author} {\bibfnamefont {A.}~\bibnamefont {Queiruga}}, \bibinfo {author} {\bibfnamefont {L.}~\bibnamefont {Hodgkinson}},\ and\ \bibinfo {author} {\bibfnamefont {M.~W.}\ \bibnamefont {Mahoney}},\ }\bibfield  {title} {\bibinfo {title} {Lipschitz recurrent neural networks},\ }in\ \href {https://openreview.net/forum?id=-N7PBXqOUJZ} {\emph {\bibinfo {booktitle} {International Conference on Learning Representations}}}\ (\bibinfo {year} {2021})\BibitemShut {NoStop}%
\bibitem [{\citenamefont {Sussillo}\ and\ \citenamefont {Abbott}(2009)}]{Sussillo:2009:Generating}%
  \BibitemOpen
  \bibfield  {author} {\bibinfo {author} {\bibfnamefont {D.}~\bibnamefont {Sussillo}}\ and\ \bibinfo {author} {\bibfnamefont {L.}~\bibnamefont {Abbott}},\ }\bibfield  {title} {\bibinfo {title} {Generating coherent patterns of activity from chaotic neural networks},\ }\href {https://doi.org/https://doi.org/10.1016/j.neuron.2009.07.018} {\bibfield  {journal} {\bibinfo  {journal} {Neuron}\ }\textbf {\bibinfo {volume} {63}},\ \bibinfo {pages} {544} (\bibinfo {year} {2009})}\BibitemShut {NoStop}%
\bibitem [{\citenamefont {Laje}\ and\ \citenamefont {Buonomano}(2013)}]{Laje:2013:Robust}%
  \BibitemOpen
  \bibfield  {author} {\bibinfo {author} {\bibfnamefont {R.}~\bibnamefont {Laje}}\ and\ \bibinfo {author} {\bibfnamefont {D.~V.}\ \bibnamefont {Buonomano}},\ }\bibfield  {title} {\bibinfo {title} {Robust timing and motor patterns by taming chaos in recurrent neural networks},\ }\href {https://doi.org/10.1038/nn.3405} {\bibfield  {journal} {\bibinfo  {journal} {Nature Neuroscience}\ }\textbf {\bibinfo {volume} {16}},\ \bibinfo {pages} {925} (\bibinfo {year} {2013})}\BibitemShut {NoStop}%
\bibitem [{\citenamefont {Chen}\ and\ \citenamefont {Aihara}(1995)}]{Chen:1995:CSA}%
  \BibitemOpen
  \bibfield  {author} {\bibinfo {author} {\bibfnamefont {L.}~\bibnamefont {Chen}}\ and\ \bibinfo {author} {\bibfnamefont {K.}~\bibnamefont {Aihara}},\ }\bibfield  {title} {\bibinfo {title} {Chaotic simulated annealing by a neural network model with transient chaos},\ }\href {https://doi.org/https://doi.org/10.1016/0893-6080(95)00033-V} {\bibfield  {journal} {\bibinfo  {journal} {Neural Networks}\ }\textbf {\bibinfo {volume} {8}},\ \bibinfo {pages} {915} (\bibinfo {year} {1995})}\BibitemShut {NoStop}%
\bibitem [{\citenamefont {Crutchfield}\ and\ \citenamefont {Kaneko}(1988)}]{crutchfield:1988:attractors}%
  \BibitemOpen
  \bibfield  {author} {\bibinfo {author} {\bibfnamefont {J.~P.}\ \bibnamefont {Crutchfield}}\ and\ \bibinfo {author} {\bibfnamefont {K.}~\bibnamefont {Kaneko}},\ }\bibfield  {title} {\bibinfo {title} {Are attractors relevant to turbulence?},\ }\href@noop {} {\bibfield  {journal} {\bibinfo  {journal} {Physical review letters}\ }\textbf {\bibinfo {volume} {60}},\ \bibinfo {pages} {2715} (\bibinfo {year} {1988})}\BibitemShut {NoStop}%
\bibitem [{\citenamefont {Lai}\ and\ \citenamefont {T{\'e}l}(2011)}]{lai:2011:transient}%
  \BibitemOpen
  \bibfield  {author} {\bibinfo {author} {\bibfnamefont {Y.-C.}\ \bibnamefont {Lai}}\ and\ \bibinfo {author} {\bibfnamefont {T.}~\bibnamefont {T{\'e}l}},\ }\href@noop {} {\emph {\bibinfo {title} {Transient chaos: complex dynamics on finite time scales}}},\ Vol.\ \bibinfo {volume} {173}\ (\bibinfo  {publisher} {Springer Science \& Business Media},\ \bibinfo {year} {2011})\BibitemShut {NoStop}%
\bibitem [{\citenamefont {Chen}\ and\ \citenamefont {Aihara}(1999)}]{Chen:1999:GlobalSearching}%
  \BibitemOpen
  \bibfield  {author} {\bibinfo {author} {\bibfnamefont {L.}~\bibnamefont {Chen}}\ and\ \bibinfo {author} {\bibfnamefont {K.}~\bibnamefont {Aihara}},\ }\bibfield  {title} {\bibinfo {title} {Global searching ability of chaotic neural networks},\ }\href {https://doi.org/10.1109/81.780378} {\bibfield  {journal} {\bibinfo  {journal} {IEEE Transactions on Circuits and Systems I: Fundamental Theory and Applications}\ }\textbf {\bibinfo {volume} {46}},\ \bibinfo {pages} {974} (\bibinfo {year} {1999})}\BibitemShut {NoStop}%
\bibitem [{\citenamefont {Herrmann}\ \emph {et~al.}(2022)\citenamefont {Herrmann}, \citenamefont {Granz},\ and\ \citenamefont {Landgraf}}]{herrmann:2022:chaotic}%
  \BibitemOpen
  \bibfield  {author} {\bibinfo {author} {\bibfnamefont {L.}~\bibnamefont {Herrmann}}, \bibinfo {author} {\bibfnamefont {M.}~\bibnamefont {Granz}},\ and\ \bibinfo {author} {\bibfnamefont {T.}~\bibnamefont {Landgraf}},\ }\bibfield  {title} {\bibinfo {title} {Chaotic dynamics are intrinsic to neural network training with {SGD}},\ }in\ \href {https://openreview.net/forum?id=ffy-h0GKZbK} {\emph {\bibinfo {booktitle} {Advances in Neural Information Processing Systems}}},\ \bibinfo {editor} {edited by\ \bibinfo {editor} {\bibfnamefont {A.~H.}\ \bibnamefont {Oh}}, \bibinfo {editor} {\bibfnamefont {A.}~\bibnamefont {Agarwal}}, \bibinfo {editor} {\bibfnamefont {D.}~\bibnamefont {Belgrave}},\ and\ \bibinfo {editor} {\bibfnamefont {K.}~\bibnamefont {Cho}}}\ (\bibinfo {year} {2022})\BibitemShut {NoStop}%
\bibitem [{\citenamefont {N~B}\ \emph {et~al.}(2019)\citenamefont {N~B}, \citenamefont {Kathpalia}, \citenamefont {Saha},\ and\ \citenamefont {Nagaraj}}]{Harikrishnan:2019:ChaosNet}%
  \BibitemOpen
  \bibfield  {author} {\bibinfo {author} {\bibfnamefont {H.}~\bibnamefont {N~B}}, \bibinfo {author} {\bibfnamefont {A.}~\bibnamefont {Kathpalia}}, \bibinfo {author} {\bibfnamefont {S.}~\bibnamefont {Saha}},\ and\ \bibinfo {author} {\bibfnamefont {N.}~\bibnamefont {Nagaraj}},\ }\bibfield  {title} {\bibinfo {title} {Chaosnet: A chaos based artificial neural network architecture for classification},\ }\href {https://doi.org/10.1063/1.5120831} {\bibfield  {journal} {\bibinfo  {journal} {Chaos: An Interdisciplinary Journal of Nonlinear Science}\ }\textbf {\bibinfo {volume} {29}},\ \bibinfo {pages} {113125} (\bibinfo {year} {2019})}\BibitemShut {NoStop}%
\bibitem [{\citenamefont {Harikrishnan}\ and\ \citenamefont {Nagaraj}(2020)}]{harikrishnan:2020:neurochaos}%
  \BibitemOpen
  \bibfield  {author} {\bibinfo {author} {\bibfnamefont {N.}~\bibnamefont {Harikrishnan}}\ and\ \bibinfo {author} {\bibfnamefont {N.}~\bibnamefont {Nagaraj}},\ }\bibfield  {title} {\bibinfo {title} {Neurochaos inspired hybrid machine learning architecture for classification},\ }in\ \href@noop {} {\emph {\bibinfo {booktitle} {2020 International Conference on Signal Processing and Communications (SPCOM)}}}\ (\bibinfo {organization} {IEEE},\ \bibinfo {year} {2020})\ pp.\ \bibinfo {pages} {1--5}\BibitemShut {NoStop}%
\bibitem [{\citenamefont {Sudeesh}\ \emph {et~al.}(2023)\citenamefont {Sudeesh}, \citenamefont {Nair}, \citenamefont {Suravajhala} \emph {et~al.}}]{sudeesh:2023:biologically}%
  \BibitemOpen
  \bibfield  {author} {\bibinfo {author} {\bibfnamefont {A.}~\bibnamefont {Sudeesh}}, \bibinfo {author} {\bibfnamefont {P.~P.}\ \bibnamefont {Nair}}, \bibinfo {author} {\bibfnamefont {P.}~\bibnamefont {Suravajhala}}, \emph {et~al.},\ }\bibfield  {title} {\bibinfo {title} {Biologically inspired chaosnet architecture for hypothetical protein classification},\ }\href@noop {} {\bibfield  {journal} {\bibinfo  {journal} {arXiv preprint arXiv:2302.02427}\ } (\bibinfo {year} {2023})}\BibitemShut {NoStop}%
\bibitem [{\citenamefont {Harikrishnan}\ \emph {et~al.}(2022)\citenamefont {Harikrishnan}, \citenamefont {Pranay},\ and\ \citenamefont {Nagaraj}}]{harikrishnan:2022:classification}%
  \BibitemOpen
  \bibfield  {author} {\bibinfo {author} {\bibfnamefont {N.}~\bibnamefont {Harikrishnan}}, \bibinfo {author} {\bibfnamefont {S.}~\bibnamefont {Pranay}},\ and\ \bibinfo {author} {\bibfnamefont {N.}~\bibnamefont {Nagaraj}},\ }\bibfield  {title} {\bibinfo {title} {Classification of sars-cov-2 viral genome sequences using neurochaos learning},\ }\href@noop {} {\bibfield  {journal} {\bibinfo  {journal} {Medical \& Biological Engineering \& Computing}\ }\textbf {\bibinfo {volume} {60}},\ \bibinfo {pages} {2245} (\bibinfo {year} {2022})}\BibitemShut {NoStop}%
\bibitem [{\citenamefont {He}\ \emph {et~al.}(2015)\citenamefont {He}, \citenamefont {Zhang}, \citenamefont {Ren},\ and\ \citenamefont {Sun}}]{he:2015:delving}%
  \BibitemOpen
  \bibfield  {author} {\bibinfo {author} {\bibfnamefont {K.}~\bibnamefont {He}}, \bibinfo {author} {\bibfnamefont {X.}~\bibnamefont {Zhang}}, \bibinfo {author} {\bibfnamefont {S.}~\bibnamefont {Ren}},\ and\ \bibinfo {author} {\bibfnamefont {J.}~\bibnamefont {Sun}},\ }\bibfield  {title} {\bibinfo {title} {Delving deep into rectifiers: Surpassing human-level performance on imagenet classification},\ }in\ \href@noop {} {\emph {\bibinfo {booktitle} {Proceedings of the IEEE international conference on computer vision}}}\ (\bibinfo {year} {2015})\ pp.\ \bibinfo {pages} {1026--1034}\BibitemShut {NoStop}%
\bibitem [{\citenamefont {Glorot}\ and\ \citenamefont {Bengio}(2010)}]{glorot:2010:understanding}%
  \BibitemOpen
  \bibfield  {author} {\bibinfo {author} {\bibfnamefont {X.}~\bibnamefont {Glorot}}\ and\ \bibinfo {author} {\bibfnamefont {Y.}~\bibnamefont {Bengio}},\ }\bibfield  {title} {\bibinfo {title} {Understanding the difficulty of training deep feedforward neural networks},\ }in\ \href@noop {} {\emph {\bibinfo {booktitle} {Proceedings of the thirteenth international conference on artificial intelligence and statistics}}}\ (\bibinfo {organization} {JMLR Workshop and Conference Proceedings},\ \bibinfo {year} {2010})\ pp.\ \bibinfo {pages} {249--256}\BibitemShut {NoStop}%
\bibitem [{\citenamefont {Luko{\v{s}}evi{\v{c}}ius}\ and\ \citenamefont {Jaeger}(2009)}]{lukovsevivcius:2009:reservoir}%
  \BibitemOpen
  \bibfield  {author} {\bibinfo {author} {\bibfnamefont {M.}~\bibnamefont {Luko{\v{s}}evi{\v{c}}ius}}\ and\ \bibinfo {author} {\bibfnamefont {H.}~\bibnamefont {Jaeger}},\ }\bibfield  {title} {\bibinfo {title} {Reservoir computing approaches to recurrent neural network training},\ }\href@noop {} {\bibfield  {journal} {\bibinfo  {journal} {Computer science review}\ }\textbf {\bibinfo {volume} {3}},\ \bibinfo {pages} {127} (\bibinfo {year} {2009})}\BibitemShut {NoStop}%
\bibitem [{\citenamefont {Tanaka}\ \emph {et~al.}(2019)\citenamefont {Tanaka}, \citenamefont {Yamane}, \citenamefont {H{\'e}roux}, \citenamefont {Nakane}, \citenamefont {Kanazawa}, \citenamefont {Takeda}, \citenamefont {Numata}, \citenamefont {Nakano},\ and\ \citenamefont {Hirose}}]{tanaka:2019:recent}%
  \BibitemOpen
  \bibfield  {author} {\bibinfo {author} {\bibfnamefont {G.}~\bibnamefont {Tanaka}}, \bibinfo {author} {\bibfnamefont {T.}~\bibnamefont {Yamane}}, \bibinfo {author} {\bibfnamefont {J.~B.}\ \bibnamefont {H{\'e}roux}}, \bibinfo {author} {\bibfnamefont {R.}~\bibnamefont {Nakane}}, \bibinfo {author} {\bibfnamefont {N.}~\bibnamefont {Kanazawa}}, \bibinfo {author} {\bibfnamefont {S.}~\bibnamefont {Takeda}}, \bibinfo {author} {\bibfnamefont {H.}~\bibnamefont {Numata}}, \bibinfo {author} {\bibfnamefont {D.}~\bibnamefont {Nakano}},\ and\ \bibinfo {author} {\bibfnamefont {A.}~\bibnamefont {Hirose}},\ }\bibfield  {title} {\bibinfo {title} {Recent advances in physical reservoir computing: A review},\ }\href@noop {} {\bibfield  {journal} {\bibinfo  {journal} {Neural Networks}\ }\textbf {\bibinfo {volume} {115}},\ \bibinfo {pages} {100} (\bibinfo {year} {2019})}\BibitemShut {NoStop}%
\bibitem [{\citenamefont {Nakajima}(2020)}]{nakajima:2020:physical}%
  \BibitemOpen
  \bibfield  {author} {\bibinfo {author} {\bibfnamefont {K.}~\bibnamefont {Nakajima}},\ }\bibfield  {title} {\bibinfo {title} {Physical reservoir computing—an introductory perspective},\ }\href@noop {} {\bibfield  {journal} {\bibinfo  {journal} {Japanese Journal of Applied Physics}\ }\textbf {\bibinfo {volume} {59}},\ \bibinfo {pages} {060501} (\bibinfo {year} {2020})}\BibitemShut {NoStop}%
\bibitem [{\citenamefont {Jaeger}\ \emph {et~al.}(2007)\citenamefont {Jaeger}, \citenamefont {Luko{\v{s}}evi{\v{c}}ius}, \citenamefont {Popovici},\ and\ \citenamefont {Siewert}}]{jaeger:2007:optimization}%
  \BibitemOpen
  \bibfield  {author} {\bibinfo {author} {\bibfnamefont {H.}~\bibnamefont {Jaeger}}, \bibinfo {author} {\bibfnamefont {M.}~\bibnamefont {Luko{\v{s}}evi{\v{c}}ius}}, \bibinfo {author} {\bibfnamefont {D.}~\bibnamefont {Popovici}},\ and\ \bibinfo {author} {\bibfnamefont {U.}~\bibnamefont {Siewert}},\ }\bibfield  {title} {\bibinfo {title} {Optimization and applications of echo state networks with leaky-integrator neurons},\ }\href@noop {} {\bibfield  {journal} {\bibinfo  {journal} {Neural networks}\ }\textbf {\bibinfo {volume} {20}},\ \bibinfo {pages} {335} (\bibinfo {year} {2007})}\BibitemShut {NoStop}%
\bibitem [{\citenamefont {Thiede}\ and\ \citenamefont {Parlitz}(2019)}]{thiede:2019:gradient}%
  \BibitemOpen
  \bibfield  {author} {\bibinfo {author} {\bibfnamefont {L.~A.}\ \bibnamefont {Thiede}}\ and\ \bibinfo {author} {\bibfnamefont {U.}~\bibnamefont {Parlitz}},\ }\bibfield  {title} {\bibinfo {title} {Gradient based hyperparameter optimization in echo state networks},\ }\href@noop {} {\bibfield  {journal} {\bibinfo  {journal} {Neural Networks}\ }\textbf {\bibinfo {volume} {115}},\ \bibinfo {pages} {23} (\bibinfo {year} {2019})}\BibitemShut {NoStop}%
\bibitem [{\citenamefont {Baldi}(1995)}]{pierre:1995:gradient}%
  \BibitemOpen
  \bibfield  {author} {\bibinfo {author} {\bibfnamefont {P.}~\bibnamefont {Baldi}},\ }\bibfield  {title} {\bibinfo {title} {Gradient descent learning algorithm overview: a general dynamical systems perspective},\ }\href {https://doi.org/10.1109/72.363438} {\bibfield  {journal} {\bibinfo  {journal} {IEEE Transactions on Neural Networks}\ }\textbf {\bibinfo {volume} {6}},\ \bibinfo {pages} {182} (\bibinfo {year} {1995})}\BibitemShut {NoStop}%
\bibitem [{\citenamefont {Erichson}\ \emph {et~al.}(2021{\natexlab{b}})\citenamefont {Erichson}, \citenamefont {Azencot}, \citenamefont {Queiruga}, \citenamefont {Hodgkinson},\ and\ \citenamefont {Mahoney}}]{sp:erichson:2021}%
  \BibitemOpen
  \bibfield  {author} {\bibinfo {author} {\bibfnamefont {N.~B.}\ \bibnamefont {Erichson}}, \bibinfo {author} {\bibfnamefont {O.}~\bibnamefont {Azencot}}, \bibinfo {author} {\bibfnamefont {A.}~\bibnamefont {Queiruga}}, \bibinfo {author} {\bibfnamefont {L.}~\bibnamefont {Hodgkinson}},\ and\ \bibinfo {author} {\bibfnamefont {M.~W.}\ \bibnamefont {Mahoney}},\ }\bibfield  {title} {\bibinfo {title} {Lipschitz recurrent neural networks},\ }in\ \href {https://openreview.net/forum?id=-N7PBXqOUJZ} {\emph {\bibinfo {booktitle} {International Conference on Learning Representations}}}\ (\bibinfo {year} {2021})\BibitemShut {NoStop}%
\bibitem [{\citenamefont {Ee}(2017)}]{sp:erinan:2017}%
  \BibitemOpen
  \bibfield  {author} {\bibinfo {author} {\bibfnamefont {W.}~\bibnamefont {Ee}},\ }\bibfield  {title} {\bibinfo {title} {A proposal on machine learning via dynamical systems},\ }\href {https://doi.org/10.1007/s40304-017-0103-z} {\bibfield  {journal} {\bibinfo  {journal} {Communications in Mathematics and Statistics}\ }\textbf {\bibinfo {volume} {5}},\ \bibinfo {pages} {1} (\bibinfo {year} {2017})}\BibitemShut {NoStop}%
\bibitem [{\citenamefont {Chang}\ \emph {et~al.}(2018)\citenamefont {Chang}, \citenamefont {Meng}, \citenamefont {Haber}, \citenamefont {Tung},\ and\ \citenamefont {Begert}}]{sp:chang:2018}%
  \BibitemOpen
  \bibfield  {author} {\bibinfo {author} {\bibfnamefont {B.}~\bibnamefont {Chang}}, \bibinfo {author} {\bibfnamefont {L.}~\bibnamefont {Meng}}, \bibinfo {author} {\bibfnamefont {E.}~\bibnamefont {Haber}}, \bibinfo {author} {\bibfnamefont {F.}~\bibnamefont {Tung}},\ and\ \bibinfo {author} {\bibfnamefont {D.}~\bibnamefont {Begert}},\ }\bibfield  {title} {\bibinfo {title} {Multi-level residual networks from dynamical systems view},\ }in\ \href {https://openreview.net/forum?id=SyJS-OgR-} {\emph {\bibinfo {booktitle} {International Conference on Learning Representations}}}\ (\bibinfo {year} {2018})\BibitemShut {NoStop}%
\end{thebibliography}%

\appendix

\section{Chaos in Deep Neural Networks}

\subsection{Background}
\label{sup:chaos_in_dnns}

As the field of machine learning has progressed, research into the area of chaos has unfolded in parallel. The unpredictable nature of chaos in neural networks, stemming largely from their sensitivity to initial conditions, can complicate the generation of reliable long-term predictions and impact reproducibility. To mitigate these challenges, massive efforts have been directed at curbing the emergence of chaos in neural networks \cite{molgedey:1992:suppressing,wang:1996:suppressing,ishihara:2005:magic,rajan:2010:stimulus,li:2013:controlling,allehiany:2021:chaos}. Strange attractors \cite{Shaw:1981strange}, a unique category of chaotic dynamical systems demonstrating converging trajectories, have been scrutinized to decode the governing principles of information flow within complex systems. In the context of deep neural networks (DNNs), extensive work has been conducted to develop deep recurrent neural networks (RNNs) that maintain stable attractors throughout training \cite{laurent:2017:a, chang:2018:antisymmetricrnn, erichson:2021:lipschitz}. Similarly, within reservoir computing (RC), methods have been devised to transform chaotic activities into stable behaviors \cite{Sussillo:2009:Generating, Laje:2013:Robust}. All these efforts aim to suppress chaos within neural networks, enhancing their convergence and robustness.

However, it is important to recognize that significant computational prowess can be harnessed from the chaotic and heterogeneous states of networks. This largely stems from their rich internal dynamics, which can lay the groundwork for complex nonlinear computations and form a bridge between the past, present, and future \cite{Shaw:1981strange}. The transiently chaotic neural network (TCNN) \cite{Chen:1995:CSA} has been proposed for approximating solutions to combinatorial optimization problems by leveraging transient chaotic dynamics \cite{crutchfield:1988:attractors,lai:2011:transient} within neural networks, thereby augmenting their capability to identify globally optimal or near-optimal solutions. Further investigations have affirmed the global searching abilities of TCNNs, demonstrating that their attractor set encompasses not just local but also global minima for widely utilized objective functions \cite{Chen:1999:GlobalSearching}. Recent research on stochastic gradient descent (SGD) has elucidated comparable benefits of chaos for convergence, highlighting that chaos is intrinsic to SGD optimization \cite{herrmann:2022:chaotic}. It was shown that chaotic dynamics consistently manifest throughout the training process, persisting even after the model has achieved convergence. 

Despite the captivating benefits that chaos introduces to the optimization of neural networks, the exploration of chaos for utilizing chaotic dynamics directly still lags behind. Research has been conducted to integrate one-dimensional chaotic maps, known as the generalized luröth series (GLS), as computation neurons for use in weakly supervised image classification tasks, revealing their potential to directly exploit chaos's non-linearity \cite{Harikrishnan:2019:ChaosNet}. Subsequent studies deploying ChaosNet across various task domains further underscored its efficacy \cite{harikrishnan:2020:neurochaos, sudeesh:2023:biologically, harikrishnan:2022:classification}. However, when compared with conventional deep neural networks, ChaosNet failed to demonstrate competitive performance consistently.

\begin{figure*}[!t]
    \begin{center}
        \includegraphics[width=0.8\textwidth]{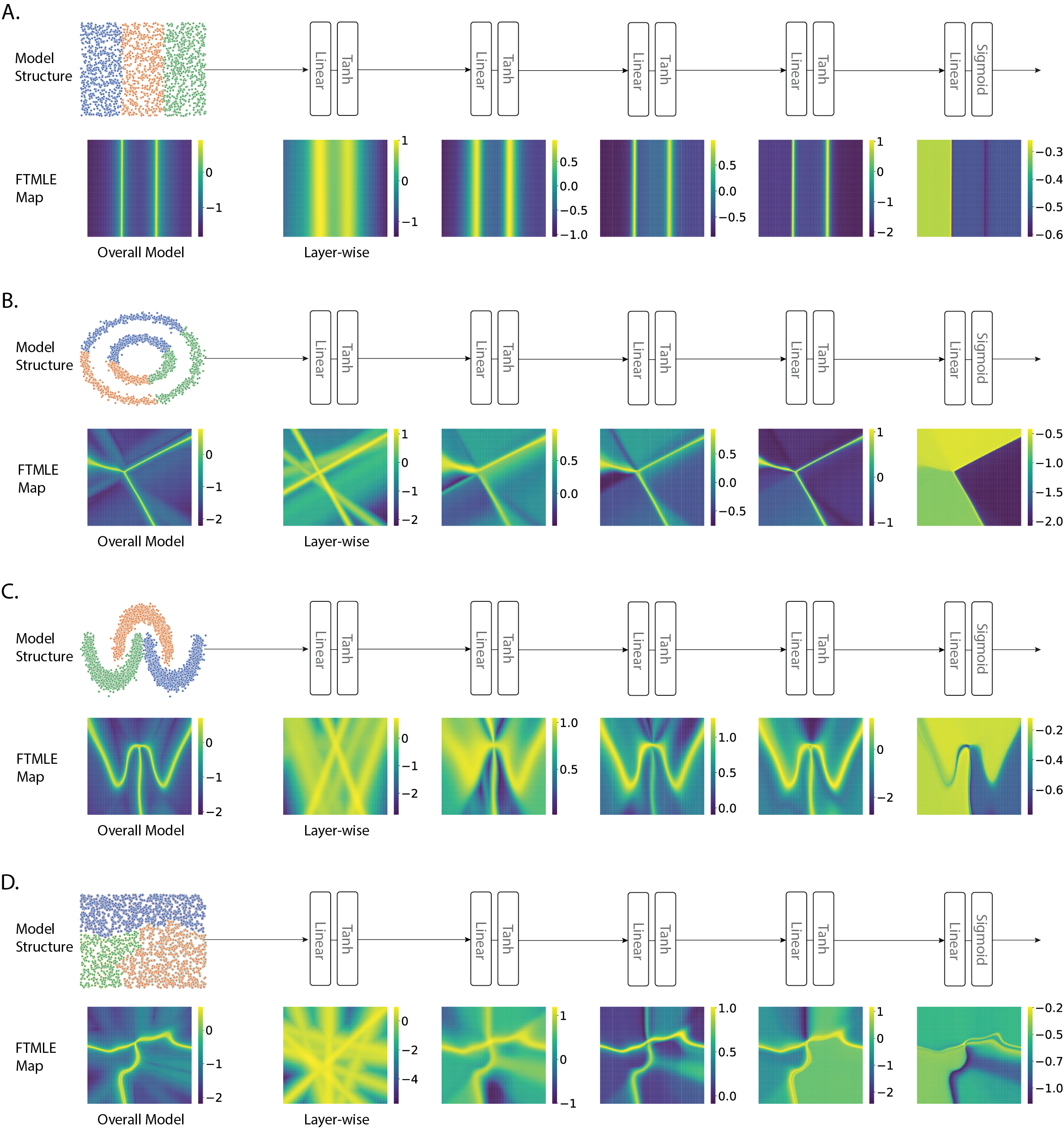}
    \end{center}
    \caption{FTMLE analyses are performed on MLPs, which have been trained using various input samples. \textbf{A--D}. The top row depicts the clusters of the input samples alongside the identical structure of the MLP model. The bottom row displays the FTMLE map computed from the testing data.}
    \label{fig:SUP-MLP-FTMLE}
\end{figure*}

A challenge in studying the dynamics of DNNs stems from their limited depth, leading these networks to exhibit transient expansion or contraction dynamics. This behavior contrasts sharply with the usual methods employed to study dynamical systems. Digging deeper, we examine the individual layers of these networks and find that they display shifts in their dimensions and exhibit diverse connection types, such as residual or upsampling links in some layers. This makes the task of understanding their dynamics using standard methods, typically suited for RNNs, quite tricky. Building on the approach of Kondo et al. \cite{Misaki:Kondo:2021}, we have turned to the finite-time maximum Lyapunov exponent (FTMLE). This represents a broader take on the maximum Lyapunov exponent (MLE) and is particularly adept at handling systems with transient dynamics. Moving beyond earlier research that centered on shallow networks, we employ this metric on the latest, deeper networks known for their standout performance across a variety of tasks. Our primary findings in the main text of this work underscore that these DNNs make use of an expansion property when processing information. This sheds light on the potential influence of chaos in separating features during this phase.

\subsection{FTMLE Analysis on MLPs}
\label{sec:sup_ftmle_mlps}

We conducted the FTMLE evaluation of MLPs that included five hidden layers. The model was trained on various two-dimensional, multi-class classification datasets. Both the overall model and the layer-wise FTMLE values were assessed. The training dataset comprised 2,000 generated samples following various distributions, and the FTMLE analysis utilized 9,000 uniformly generated test samples. Given the two-dimensional nature of the input data, we calculated the FTMLE value for each individual test sample. This allowed for a straightforward presentation of the FTMLE map, where the axes represent feature values and the color indicates the FTMLE value.

\begin{figure*}[!t]
    \begin{center}
        \includegraphics[width=0.8\textwidth]{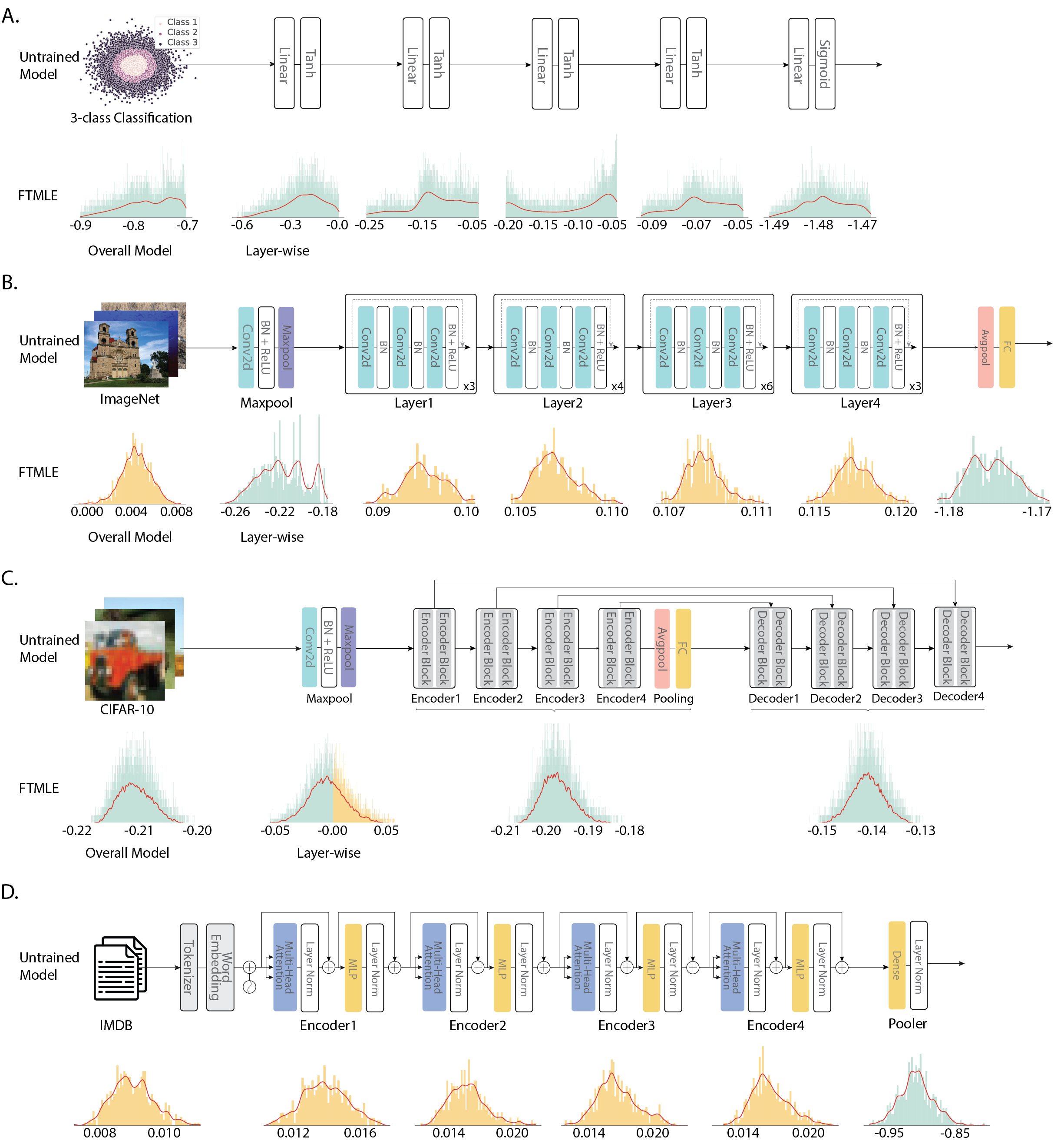}
    \end{center}
    \caption{FTMLE studies are conducted on untrained DNNs. These analyses show the FTMLE distribution, revealing both overall and layer-specific FTMLE values in response to various input samples. In the provided figures, blue regions indicate areas of contraction, whereas orange regions highlight areas of expansion. The linear layers of DNNs are initialized by a uniform distribution, and convolutional layers are initialized by the Kaiming distribution. \textbf{A.} Analysis of a untrained five-layer MLP. \textbf{B.} FTMLE study of an untrained ResNet50 model, applied to a portion of the ImageNet validation set, selecting one image from each of the 1,000 categories. \textbf{C.} Evaluation of an untrained ResNet-based auto-encoder on the CIFAR10 test dataset. \textbf{D.} Assessment of an untrained BERT-mini model on the IMDb test dataset.}
    \label{fig:SUP-UNTRAINED-DNN}
\end{figure*}

As shown in Figure \ref{fig:SUP-MLP-FTMLE}A--D, we observed that the overall model FTMLE map clearly highlights the classification boundaries, with high FTMLE values closely aligning with these boundaries. In the layer-wise FTMLE maps, we noticed the evolution of these boundaries, from a coarse overlap of straight lines to more precise non-linear boundaries. Furthermore, the first four layers exhibited an expansion behavior, indicated by the emergence of positive FTMLE values. This reveals the role of expansion behavior in the information processing of MLPs through layer propagation.

\begin{table*}[!t]
  \centering
  \small
  \resizebox{0.8\textwidth}{!}{%
  \begin{tabular}{lrrrrrrr}
    \toprule
    MLP & Overall Model &  & Layer1 & Layer2 & Layer3 & Layer4 & Layer5 \\
    \midrule
    Trained & $-1.336_{\pm.618}$ &  & $-0.016_{\pm.503}$ & $0.615_{\pm.726}$ & $0.546_{\pm.459}$  & $-2.445_{\pm1.23}$ & $-0.600_{\pm.206}$ \\
    Default & $-0.773_{\pm.047}$ &  & $-0.239_{\pm.121}$ & $-0.119_{\pm.057}$ & $-0.112_{\pm.058}$ & $-0.069_{\pm.011}$ & $-1.479_{\pm.005}$ \\
    Xavier  &  $-0.460_{\pm.098}$ &  & $0.180_{\pm.137}$ & $0.306_{\pm.105}$ & $0.422_{\pm.082}$ & $0.344_{\pm.124}$ & $-0.973_{\pm.063}$ \\
    \midrule
    ResNet-50 & Overall Model & Maxpool & Layer1 & Layer2 & Layer3 & Layer4 & Avgpool \\
    \midrule
    Trained & $0.054_{\pm.007}$  & $0.581_{\pm.001}$  & $0.317_{\pm.055}$  & $0.208_{\pm.018}$  & $0.052_{\pm.011}$ & $0.356_{\pm.027}$ & $-0.320_{\pm.040}$ \\
    Default & $0.005_{\pm.001}$  & $-0.223_{\pm.022}$ & $0.097_{\pm.003}$  & $0.107_{\pm.001}$  & $0.108_{\pm.001}$ & $0.118_{\pm.001}$ & $-1.175_{\pm.004}$ \\
    Xavier  & $-0.090_{\pm.001}$ & $-0.250_{\pm.025}$ & $-0.018_{\pm.001}$ & $-0.004_{\pm.001}$ & $-0.006_{\pm.001}$ & $-0.002_{\pm.001}$ & $-1.168_{\pm.005}$ \\
    \midrule
    Auto-encoder & Overall Model & \multicolumn{2}{r}{Maxpool} &  \multicolumn{2}{r}{Encoder} & \multicolumn{2}{r}{Decoder} \\
    \midrule
    Trained & $0.152_{\pm.013}$  & \multicolumn{2}{r}{$0.552_{\pm.008}$}  & \multicolumn{2}{r}{$-0.093_{\pm.015}$} & \multicolumn{2}{r}{$0.534_{\pm.007}$}  \\
    Default & $-0.210_{\pm.002}$ & \multicolumn{2}{r}{$-0.006_{\pm.019}$} & \multicolumn{2}{r}{$-0.197_{\pm.004}$} & \multicolumn{2}{r}{$-0.142_{\pm.003}$} \\
    Xavier  & $-0.081_{\pm.002}$ & \multicolumn{2}{r}{$-0.368_{\pm.015}$} & \multicolumn{2}{r}{$-0.109_{\pm.003}$} & \multicolumn{2}{r}{$0.039_{\pm.004}$}  \\
    \midrule
    BERT-mini & Overall Model &  & Encoder1 & Encoder2 & Encoder3 & Encoder4 & Pooler \\
    \midrule
    Trained & $0.089_{\pm.027}$ &  & $0.409_{\pm.054}$ & $0.480_{\pm.072}$ & $0.531_{\pm.057}$ & $0.583_{\pm.037}$ & $0.439_{\pm.075}$ \\
    Default & $0.009_{\pm.001}$ &  & $0.013_{\pm.002}$ & $0.016_{\pm.002}$ & $0.017_{\pm.002}$ & $0.017_{\pm.001}$ & $-0.911_{\pm.022}$ \\
    Xavier  & $0.004_{\pm.010}$ &  & $0.128_{\pm.033}$ & $0.061_{\pm.038}$ & $0.004_{\pm.025}$ & $-0.017_{\pm.008}$ & $0.076_{\pm.026}$ \\
    \bottomrule
  \end{tabular}
  }
  \caption{Comparative analysis of FTMLE values in trained versus untrained models. The table shows a side-by-side comparison of the FTMLE values for the pre-trained models and for untrained models. The columns present the overall model or layer-wise FTMLE values, respectively. For the untrained models, weight initialization is conducted using either the default method or the Xavier initialization. The values are presented as the mean and standard deviation of the FTMLE distribution, examined through the lens of the overall model as well as its sub-layers. The input samples for both the trained and untrained models are the same. The mean FTMLE values are computed by averaging the FTMLE values across all these individual input samples. In the untrained models, the bias value in linear layers is initialized as 0. Similarly, in the LayerNorm layers of BERT-mini, the weights ($\gamma$) become 1, while the bias ($\beta$) is set as 0.}
  \label{tab:dnn_ftmle_avg}
\end{table*}

\begin{table*}[!t]
  \centering
  \scriptsize
  \resizebox{0.8\textwidth}{!}{%
  \begin{tabular}{lrrrrrrr}
    \toprule
    MLP & Overall Model &  & Layer1 & Layer2 & Layer3 & Layer4 & Layer5 \\
    \midrule
    Trained & 0.717  &  & 0.796  & 1.490  & 1.541  & 1.471  & 0.072  \\
    Default & -0.701 &  & -0.028 & -0.027 & -0.042 & -0.051 & -1.469 \\
    Xavier  & -0.165 &  & 0.510  & 0.484  & 0.622  & 0.576  & -0.810 \\
    \midrule
    ResNet-50 & Overall Model & Maxpool & Layer1 & Layer2 & Layer3 & Layer4 & Avgpool \\
    \midrule
    Trained & 0.079  & 0.584  & 0.424  & 0.263  & 0.079  & 0.435  & -0.214  \\
    Default & 0.009  & -0.180 & 0.110  & 0.112  & 0.111  & 0.121  & -1.158 \\
    Xavier  & -0.087 & -0.212 & -0.013 & -0.003 & -0.006 & -0.001 & -0.092 \\
    \midrule
    Auto-encoder & Overall Model & \multicolumn{2}{r}{Maxpool} & \multicolumn{2}{r}{Encoder} & \multicolumn{2}{r}{Decoder} \\
    \midrule
    Trained & 0.250  & \multicolumn{2}{r}{0.595}  & \multicolumn{2}{r}{0.024}   & \multicolumn{2}{r}{0.581}  \\
    Default & -0.201 & \multicolumn{2}{r}{0.076}  & \multicolumn{2}{r}{-0.182}  & \multicolumn{2}{r}{-0.131} \\
    Xavier  & -0.075 & \multicolumn{2}{r}{-0.302} & \multicolumn{2}{r}{-0.095}  & \multicolumn{2}{r}{0.054}  \\
    \midrule
    BERT-mini & Overall Model &  & Encoder1 & Encoder2 & Encoder3 & Encoder4 & Pooler \\
    \midrule
    Trained & 0.167 &  & 0.572 & 0.652 & 0.695 & 0.735 & 0.796 \\
    Default & 0.012 &  & 0.019 & 0.021 & 0.024 & 0.022 & -0.822 \\
    Xavier  & 0.020 &  & 0.170 & 0.143 & 0.063 & 0.001 & 0.177 \\
    \bottomrule
  \end{tabular}
  }
  \caption{Comparative analysis of maximum FTMLE values in trained and untrained models. This table presents the maximum FTMLE values in the FTMLE distributions computed in Table~\ref{tab:dnn_ftmle_avg} as complementary information. The columns present the overall model or layer-wise FTMLE values, respectively.}
  \label{tab:dnn_ftmle_max}
\end{table*}

\subsection{FTMLE Analysis on Untrained DNNs}
\label{sup:untrained_dnns}

To investigate whether the expansion behavior in DNNs arises from training, we carried out FTMLE analysis on untrained DNNs. We examined the FTMLE distributions for identical input samples in DNNs loaded with pre-trained weights and those with randomly initialized weights. We employed two distinct initializations. The first was the standard one used before pretraining, which is a uniform distribution in the range $[-\textit{bound}, \textit{bound}]$ for linear layers, where $\textit{bound} = \frac{1}{\sqrt{\textit{fan\_in}}}$, and $\textit{fan\_in}$ represents the input dimension of the layer. For convolutional layers (employed in both ResNet-50- and ResNet-18-based auto-encoders), the model used Kaiming initialization \cite{he:2015:delving}, a normal distribution with a mean of 0 and a standard deviation of $\frac{2}{\sqrt{\textit{fan\_in}}}$. In BERT-mini, the weights were initialized using a normal distribution with a mean of $0$ and a standard deviation of $0.02$. The alternative approach we used was the Xavier initialization \cite{glorot:2010:understanding}. Figure \ref{fig:SUP-UNTRAINED-DNN} displays the FTMLE distributions for untrained DNNs with default weight initialization; expansion and contraction areas are highlighted in orange and blue, respectively. Observations from Figure \ref{fig:SUP-UNTRAINED-DNN}B--D indicate that the randomly initialized model and its sub-layers had regions with positive FTMLE values. For instance, in the case of ResNet-50 (Figure \ref{fig:SUP-UNTRAINED-DNN}B), the model predominantly exhibited a positive FTMLE distribution, with the four bottleneck layers also demonstrating closely aligned positive FTMLE values.

Table~\ref{tab:dnn_ftmle_avg} compares the FTMLE analysis results for pre-trained versus untrained DNNs. We noticed that trained models typically showed higher FTMLE values compared to untrained ones. Furthermore, trained models demonstrated a greater standard deviation in their FTMLE distribution. This suggests that trained models experienced more varied dynamics across different input samples, with some showing high positive FTMLE values and others displaying low negative values. This variance implies varying contributions to the training process depending on the input samples. Inputs with high FTMLE values may contribute more significantly to feature separation in the current model, a factor that changes as the model's weights are updated during training. Table~\ref{tab:dnn_ftmle_max} provides the maximum FTMLE values, complementing the mean values with additional context. Although the mean FTMLE values for the MLP model indicate more negative values compared to the two variants with random initialization, the maximum value table reveals that the trained MLP model exhibits positive FTMLE values, unlike its untrained counterparts. This leads to the conclusion that the expansion behavior observed in pre-trained DNNs is a result of the training process, in which the expansion property plays a crucial role in information processing.

\subsection{FTMLE Analysis on BERT-mini}
\label{sec:sup_bert_ftmle}

\begin{figure*}[!t]
    \begin{center}
        \includegraphics[width=0.8\textwidth]{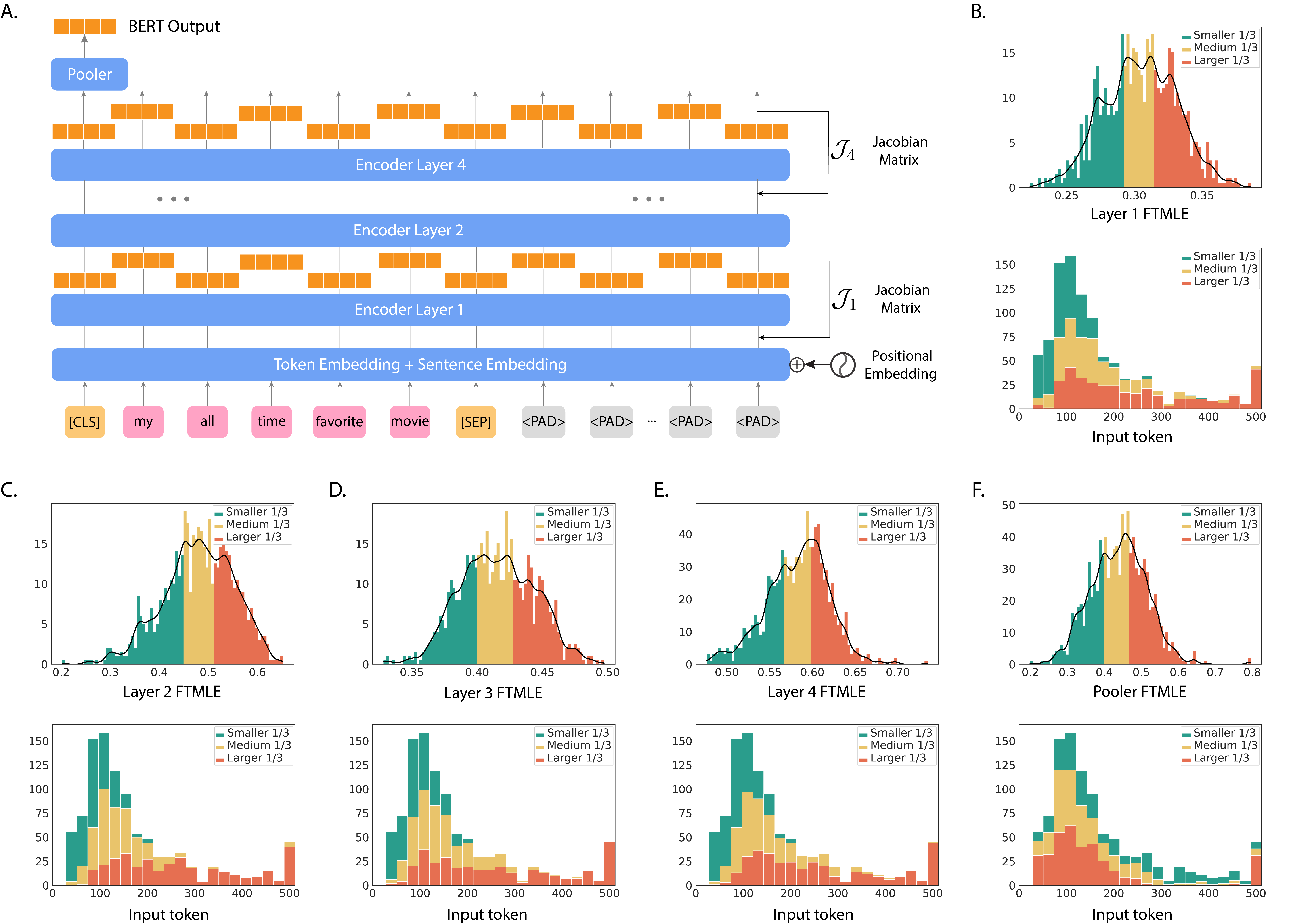}
    \end{center}
    \caption{FTMLE analysis on BERT-mini. \textbf{A.} Visualization of the BERT-mini model architecture. Input tokens (derived from tokenized input sentences) and padding tokens are highlighted in pink and gray, respectively. \textbf{B--F.} The upper figures illustrate FTMLE values distributed across three partitions (lower one-third, middle one-third, and upper one-third). The horizontal axis indicates the FTMLE value. The bottom figures represent the input token counts for each input instance, with colors indicating their respective FTMLE partitions. The horizontal axis is the count of input tokens. The maximum count of tokens that BERT-mini can process is 512.}
    \label{fig:SUP-BERT}
\end{figure*}

Prior research \cite{Misaki:Kondo:2021} has elucidated a correlation between the distribution of the FTMLE and the decision boundary within classification tasks, wherein regions with high FTMLE values tend to congregate near the decision boundary. However, the interpretability of the FTMLE remains obscured for other task types, such as regression tasks. Motivated by this gap, we embarked on preliminary investigations within the domain of natural language processing (NLP), examining the representation of text embedding in terms of the FTMLE. The outcomes of these exploratory studies have the potential to illuminate new aspects of the FTMLE and contribute to our understanding of its roles in a broader view beyond classification.

In the area of NLP, BERT \cite{devlin:2018:bert} has consistently been a cornerstone. To attain a more comprehensive understanding of the contextual surroundings, BERT employs a bidirectional training strategy, simultaneously processing linguistic entities from both preceding and succeeding directions. The architectural composition of BERT, consisting of a stack of numerous transformer encoder layers capable of task-dependent fine-tuning, renders it a remarkably versatile instrument for a broad range of natural language processing applications, such as text classification and entity recognition. By considering each sub-layer of the model as a discrete timestep within a dynamical system, we could compute the FTMLE with respect to sub-layers. This offers us the opportunity to interpret models' encoding procedures through the lens of dynamical systems theory, allowing us to uncover correlations between expansion or contraction behaviors and the information ``richness'' of individual sample instances.

Since BERT is constituted by a stacked arrangement of transformer encoder layers, the dimensions of the encoder's input and output remain consistent. This consistent dimensionality negates the effects of explicit feature space reduction, thereby facilitating a direct evaluation of the expansion or contraction behavior observed during propagation through layers. Given the substantial computation efforts dedicated to the FTMLE analysis of the standard BERT model, we have opted to employ BERT-mini, an compact BERT variant characterized by four encoder layers and a latent embedding size of 256. We evaluated BERT-mini on the IMDb benchmark \cite{maas:2011:learning}, a well-established corpus in NLP sentiment analysis, consisting of a significant number of movie reviews and balanced token length. In the context of BERT's fixed token bandwidth and implicit padding addition during tokenization, an intuitive quantifiable measure of an input's information ``richness'' can be derived from its number of input (non-padding) tokens. To analyze the distribution of FTMLE values across individual sub-layers, we delineated three FTMLE sub-regions based on their magnitude: these encompass values within the lower one-third, the middle one-third, and the upper one-third ranges, while concurrently tracking the associated number of input tokens. 

Figure \ref{fig:SUP-BERT}A shows the model architecture and the input tokens and padding tokens are shaded in pink and gray, respectively. The FTMLE values were calculated in a layer-wise manner by deriving the maximum singular values among the layer-wise Jacobian matrix. As illustrated in Figure \ref{fig:SUP-BERT}B--F, the upper figures present the distribution of FTMLE values colored according to the three different partitions, in which the bottom figures display the input token distribution shaded by the corresponding mapping of the FTMLE partitions. We observed that the four encoder layers exhibit a pattern in which instances with a larger number of input tokens exhibit more expansion compared with instances with fewer input tokens. This outcome may arise due to the encoding procedure of rich instances (those with a greater number of input tokens), necessitating a larger expansion behavior to faithfully retain the information within the encoded representation. Conversely, in the pooling layer, where the model further processed information corresponding to the first token (typically the \textit{[CLS]} token), the previously observed association was less evident.

\section{Supplementary Methods}

\subsection{Reservoir Computing with Trainable Read-in Weight}

RC stands out as a potent machine learning technique \cite{maass:2002:real,jaeger:2004:harnessing,lukovsevivcius:2009:reservoir,nakajima:2021:reservoir}, offering a compelling approach to fine-tuning neuromorphic systems and expediting a diverse array of computation tasks \cite{tanaka:2019:recent,nakajima:2020:physical}. This method harnesses the intrinsic dynamics of the system, creating what is known as a ``reservoir'' of internal nonlinear function that processes input data in a unique manner. An RC system, which is based on an RNN, comprises a reservoir that maps inputs to a high-dimensional reservoir space, and a readout layer captures patterns from the reservoir states. In contrast to the conventional learning procedures based on back-propagation through time (BPTT), the reservoir remains fixed, and only the readout weight is tuned. This highlights the primary advantage of RC, which is its capacity for rapid learning and low computational demand. In addition, the internal reservoir without adaptive updating reveals potential for hardware implementation using a variety of physical systems, substrates, and devices \cite{nakajima:2020:physical}. By incorporating a differentiable reservoir, the system can introduce extra tunable read-in layers that can more effectively map the input space to reservoir states \cite{jaeger:2007:optimization, thiede:2019:gradient}.

\subsection{Dynamical System View of Neural Networks}

In the context of neural networks, the process of gradient descent (GD) is aptly viewed through the lens of dynamical systems. This perspective emerges from the evolving nature of the network's weights and biases over time, effectively delineating a trajectory that describes the state of the system \cite{pierre:1995:gradient,herrmann:2022:chaotic}. Conceptually, the neural network's loss function delineates a space within a high-dimensional parameter landscape. Within this space, the progression of GD is comparable to a trajectory descending along this landscape, effectively mirroring the dynamics of the training stages.

Moreover, the process of information propagation during the inference phase of a neural network constitutes a dynamical system, wherein each sub-layer is representative of a timestamp. This perspective finds particularly salient expressions in RNNs, which can be analytically depicted as continuous-time dynamical systems \cite{sp:erichson:2021}. Similarly, the characterization of residual networks as ordinary differential equations—a specific category of dynamical systems—offers further evidence of the deep connection between neural networks and dynamical systems \cite{sp:erinan:2017, sp:chang:2018}.

In this paper, we expand on the approach of conceptualizing a neural network as a dynamical system. We adopt a general perspective on any form of feed-forward neural network, delving into its internal dynamics during information propagation. To accomplish this, we leverage the FTMLE, using it as a probe to inspect the internal dynamics of the sub-layers within a deep neural network. This extension of viewing the neural network as a dynamical system offers novel insights and avenues for understanding the intricate interactions within the network and their implications for information propagation and network performance.

\subsection{Adjoint Sensitivity Method for Training Continuous-Time Systems}
To train the linear read-in layer $W_{\rm in}$ for continuous-time dynamical systems, we employed the adjoint sensitivity method \cite{pontryagin:1987:mathematical}. It is an approach used to compute the gradient or sensitivity of an objective function with respect to a large set of parameters, especially when direct differentiation is computationally expensive. Consider a general system governed by the following differential equation:
\begin{equation}
\frac{du(t, p)}{dt} = f(u(t, p), p) ~,
\end{equation}
where $u(t, p)$ is the state variable, $t$ is time, and $p$ is a set of parameters. Given an objective function $J(u, p)$, we were interested in computing its gradient with respect to the parameters $p$: $\frac{dJ}{dp}$. The adjoint sensitivity method proceeds according to the following steps
\begin{enumerate}
    \item \textbf{Direct Problem}: Solve the original differential equation for $u(t, p)$ given an initial condition and $p$.

    \item \textbf{Adjoint Problem}: Define an adjoint variable $\lambda(t)$ and solve the associated adjoint equation, typically derived from the direct problem and the objective function:
    \begin{equation}
    \frac{d\lambda(t)}{dt} = -\lambda(t)^T \frac{\partial f}{\partial u} ~.
    \end{equation}
    The adjoint equation is solved backward in time, from an ending time $T$ to the initial time.

    \item \textbf{Gradient Computation}: The gradient of the objective function with respect to the parameter $p$ is calculated as:
    \begin{equation}
    \frac{dJ}{dp} = \int_0^T \lambda(t)^T \frac{\partial f}{\partial p} dt ~.
    \end{equation}
\end{enumerate}

Within the deep learning landscape, the adjoint sensitivity method plays a crucial role in tackling neural ordinary differential equations (ODEs) \cite{chen:2018:advances}. Traditional DNNs use discrete layers, whereas neural ODEs model the network's hidden states continuously over time. Training these models requires efficient gradient computation, which, if done directly through the ODE solver, is computationally intensive. The adjoint sensitivity method addresses this by allowing efficient gradient calculations. In our experiments, we used {\it PyTorch}, which utilized the adjoint sensitivity method to back-propagate the gradient of the Lorenz 96 and coupled STOs systems to train the linear read-in layer $W_{\rm in}$.

\section{Supplementary Experiments}

\subsection{Accuracy of Proposed Framework on MNIST Dataset}
\label{sec:sup_framework_eq}

In Table 1 of the main text, we demonstrate the superior performance of our frameworks on the MNIST dataset in comparison to both MLP and CNN with the same number of neurons. Specifically for FFESN, we selected models demonstrating three distinct dynamics: $\rho=0.9$ representing fix-point-convergence, $\rho=1.0$ for periodic dynamics, and $\rho=1.8$ for chaotic dynamics, each chosen for optimal performance at a certain iteration time $T$. We observed that FFESNs that showed expansion property ($\rho=1.8$) and exhibited higher accuracy against their counterparts. As for Lorenz 96 and coupled STOs, which consistently showed expansion behavior, we selected models from a range of distinctive settings that present optimal accuracy.

In addition, we extended our framework to incorporate a concatenated deep architecture in coupled STOs using the following definition:

\begin{eqnarray}
{\bf x}_1(0) &=& f_{\rm norm}\left(W_{\rm in} {\bf u}(n)\right) ~, \\
{\bf x}_k(t_k) &=& {\bf x}_k(0) + \int_{0}^{t_k} {\dot{{\bf x}_k}(s)ds}  \;\; (k=1, \ldots, M) \label{deepContinuousDNN} ~, \\
{\bf x}_{k + 1} (t) &=& W_k {\bf x}_k \;\; (k=1, \ldots, M-1) ~,~\text{and}\\
{\bf y}(n) &=& W_{\rm out} [ {\bf x}_{M}(T)^{\top} ; 1]^{\top} ~,
\end{eqnarray}

\noindent where internal weight $W_k$ is tunable, and $f_{\rm norm}$ is a normalizing function introduced in the Methods section. The deep architecture exhibited higher accuracy against a single coupled-STOs-system, showing potential for building advanced deep architecture for optimal performance. Furthermore, by integrating a tunable convolutional layer before the coupled STOs in our framework, the formulation can be described as follows

\begin{eqnarray}
\tilde{{\bf u}}(n) &=& f_{\rm conv} {\bf u}(n) ~\text{and} \\
{\bf x}_1(0) &=& f_{\rm norm}\left(W_{\rm in} \tilde{{\bf u}}(n)\right) ~,
\end{eqnarray}

\begin{figure*}[!t]
    \begin{center}
        \includegraphics[width=0.8\textwidth]{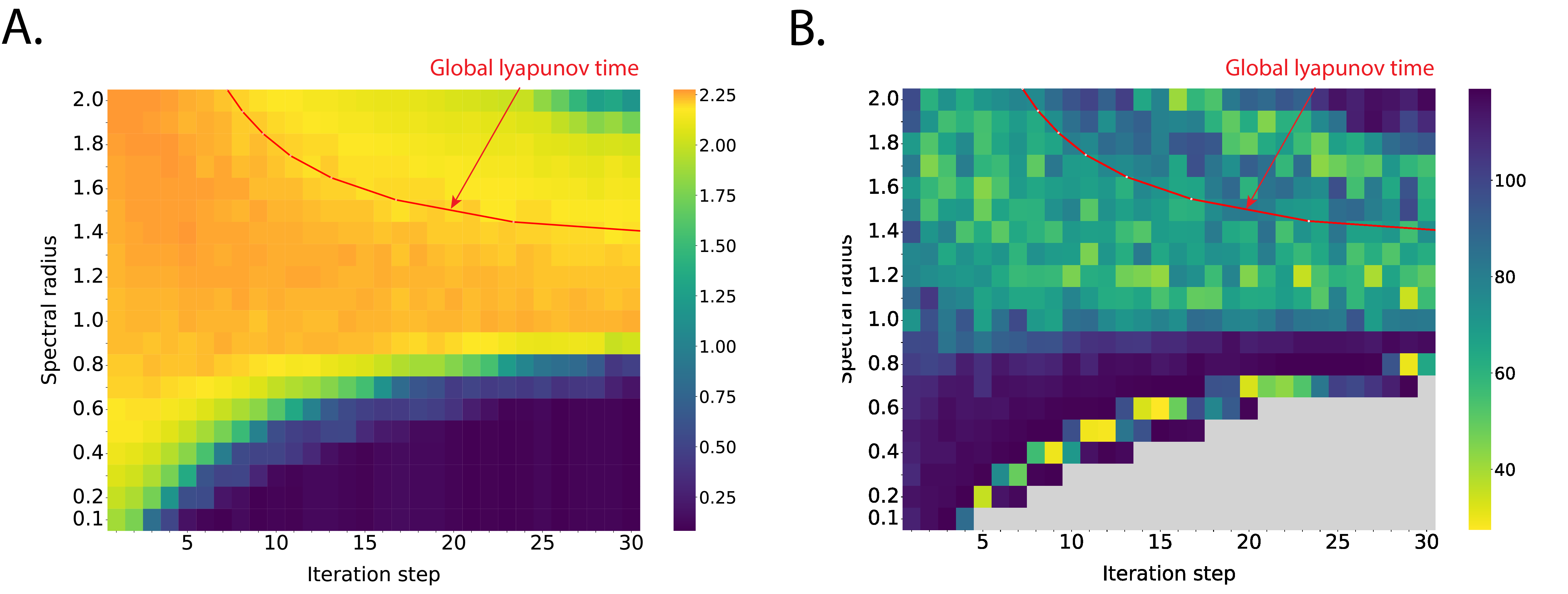}
    \end{center}
    \caption{Assessment of FFESN performance on Fashion-MNIST. The reported results represent the average of five measurements. \textbf{A.} The accuracy of the FFESN on Fashion-MNIST. The color metric represents $\left| \log(\epsilon) \right|$, where $\epsilon$ is the prediction error. The red line indicates the global Lyapunov time. \textbf{B.} The heat map represents the converging speed of the FFESN. The color metric mirrors the number of epochs required to achieve peak accuracy (with a $5e-4$ error margin). The gray region indicates the divergent zones where the FFESN fails to converge. Occasionally, lower epoch values emerge within the non-chaotic region ($\rho < 1$), a case attributed to early over-fitting.}
    \label{fig:SUP-FMNIST}
\end{figure*}

\noindent where $\tilde{{\bf u}}(n)$ represents the input post-convolutional layer processing, and $f_{\rm conv}$ is the convolution function. This configuration of coupled STOs with an initial convolutional input layer demonstrated superior accuracy compared to a CNN model with a single convolutional layer and even outperformed state-of-the-art neuromorphic STOs models in image classification \cite{leroux:2022:convolutional}.

\subsection{Feed-Forward ESN on Fashion-MNIST Dataset}
\label{sec:sup_ffesn_fmnist}

As a way of verifying the applicability of our conclusions regarding the FFESN on MNIST, we applied the FFESN to Fashion-MNIST, a comparably more challenging dataset. We trained the FFESN under identical conditions, with the spectral radius $\rho$ varying from $0.1$ to $2.0$, and iteration steps $T$ from $1$ to $30$, as in the MNIST settings. The maximum training epoch was extended to $120$. As illustrated in Figure \ref{fig:SUP-FMNIST}A, Fashion-MNIST displayed a comparable pattern, in which the chaotic region ($\rho > 1$) showed higher accuracy than the non-chaotic region ($\rho <= 1$). Additionally, the global Lyapunov time (GLT) corresponded with the zone of optimal performance. In the non-chaotic region, due to contraction behavior, the input features tended to stabilize at a constant value, leading to a rapid loss of separation capability after certain iteration steps. Conversely, in the chaotic FFESN, characterized by expansion, performance remained stable for longer, and an optimal region is identified, delineated by GLT, where the system became extensively chaotic. Figure \ref{fig:SUP-FMNIST}B demonstrates the training epochs needed to achieve convergence (optimal accuracy within an error tolerance of $5e-4$). Excluding areas showing non-convergence and adjacent regions of early overfitting, the region converging quickly (requiring fewer training epochs) also partially coincided with the optimal accuracy region. The findings from Fashion-MNIST suggest that the advantages of chaotic FFESN, such as optimal accuracy and rapid convergence, are consistent across different tasks.

\subsection{Initializing the Training of MLPs with Chaotic Weights}
\label{sec:sup_mlp_init}

\begin{figure}[!t]
    \centering
        \hspace*{-1cm}
        \includegraphics[width=10cm]{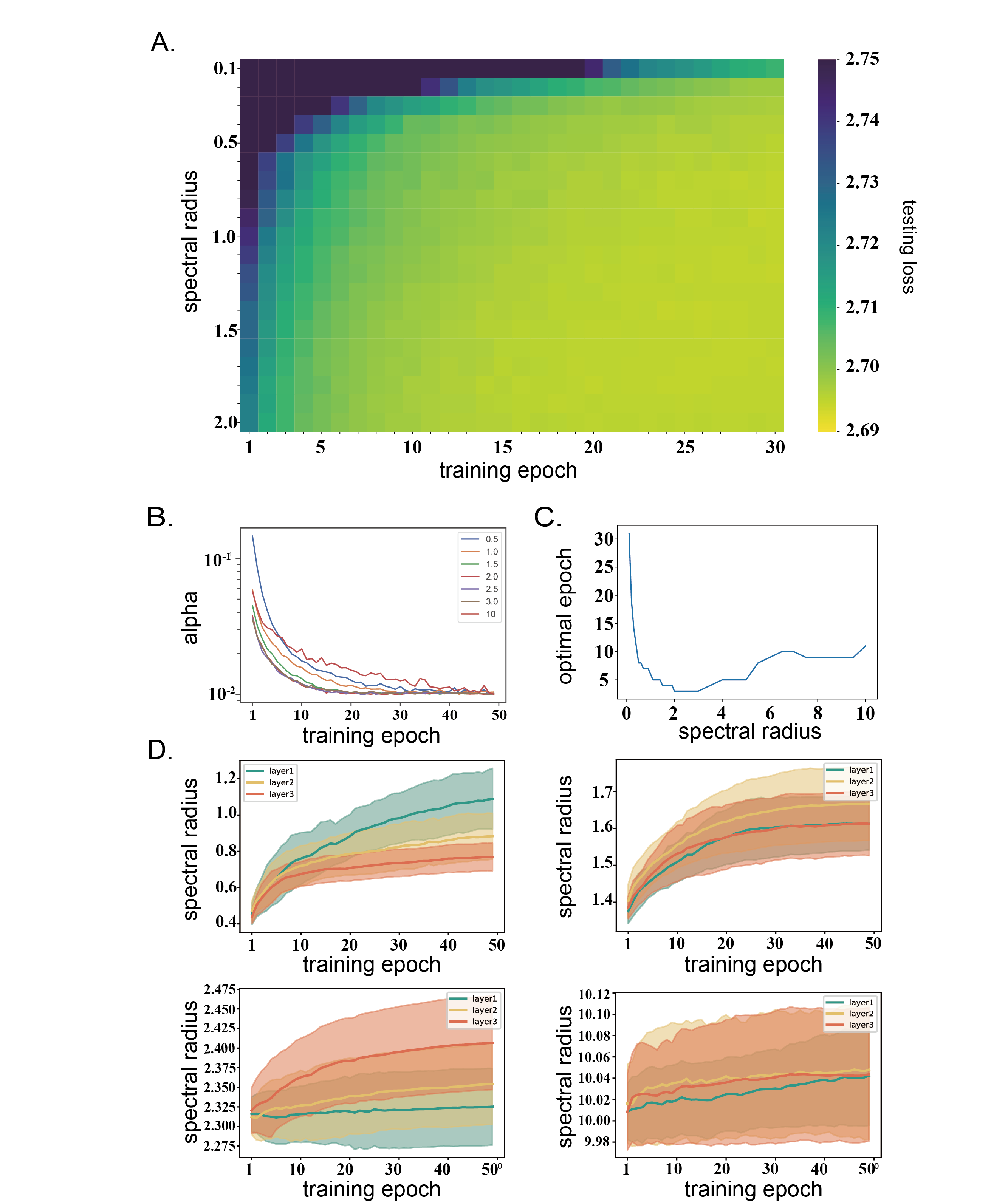}
    \caption{The illustrations of the converging speed for MLPs with different initializing weights governed by the spectral radius $\rho$. All diagrams present averaged results of ten trials. \textbf{A.} The heat map of testing loss of different $\rho$ initialization against training epochs on the MNIST dataset. The color indicates the testing loss. \textbf{B.} The tendency of testing loss versus training epochs. \textbf{C.} The relationship of the optimal training epoch and $\rho$ initialization. The converging speed is measured by the optimal training epoch, which stands for the number of epoch required to reach optimal testing loss. \textbf{D.} The spectral radius of the hidden states of MLPs during the training. We chose MLPs initialized with $\rho$ of 0.3, 1.3, 2.3 and an extreme case 10, and track the changes of $\rho$ during training. The standard deviation is shaded in colored regions.}
    \label{fig:SUP-MLP}
\end{figure}

In the main text, we delved into the impact of varying spectral radius ($\rho$) on the converging speed during training within an FFESN. Given that the internal weights of the ESN or FFESN remained fixed throughout training (kept constant to preserve its inherent dynamics), we extended our investigation to MLPs as a broader case study. Specifically, we employed MLPs comprising three hidden layers, each with the same input and output dimensions. This configuration allowed us to compute and track their weights using $\rho$. The weights of the hidden layers were initialized with a specific $\rho$ value and were subject to updates during optimization. We monitored the $\rho$ values of the weights throughout training to assess changes in dynamics. In the experiments, we employed SGD with a learning rate of 1e-3, MLPs with hidden state of 256. To illustrate the relationship between $\rho$ and convergence speed, we recorded the number of training epochs required to reach the minimum testing loss with a specific threshold. This threshold is defined as:

\begin{equation}
\alpha_{\rm t} = \frac{loss_{\rm t} - loss_{\rm min}}{loss_{\rm min}} ~,
\end{equation}

\noindent where $loss_{\rm min}$ represents the minimal testing loss obtained during training, and $loss_{\rm t}$ denotes the testing loss of the current epoch $t$. We consider $\alpha \leq 0.01$ as indicative of training converged. A lower number of training epochs signifies faster convergence.

As shown in Figure \ref{fig:SUP-MLP}A, the heat map illustrates the testing loss derived from MLPs initialized with varying $\rho$ values. It was evident that MLPs initialized with chaotic weights, such as $\rho > 1.0$, exhibited a more rapid decay in testing loss and achieved lower loss compared to their non-chaotic counterparts. Specifically, Figure \ref{fig:SUP-MLP}B presents the trend of $\alpha$ across training epochs, including additional results for $\rho > 2.0$. The result is depicted in Figure \ref{fig:SUP-MLP}C, where the regions of faster convergence are located within specific chaotic ranges of $\rho$, such as $2 < \rho < 3$, resulting in faster convergence speeds. Figure \ref{fig:SUP-MLP}D displays the alterations in $\rho$ across three hidden layers for MLPs. The four plots depict MLPs initialized with $\rho$ values of 0.3, 1.3, 2.3, and 10 respectively. We noticed that across all configurations, the $\rho$ values of all three hidden layers tended to increase during training. In an extreme scenario in which the MLP was initialized with $\rho=10$, the $\rho$ values of hidden weights presented minor changes.

The expanded experiments involving the initialization of MLPs with different $\rho$ values indicated that chaotic weights played a role in accelerating training convergence. Although our empirical findings are constrained to specific settings and lack theoretical analysis, they still underscore the efficacy of chaos in enhancing model convergence, extending the observations from FFESN to MLPs.

\subsection{Average FTMLEs in Coupled Spin-Torque Oscillators}
\label{sec:sup_csto_ftmle}

\begin{figure}[!t]
    \begin{center}
        \includegraphics[width=8cm]{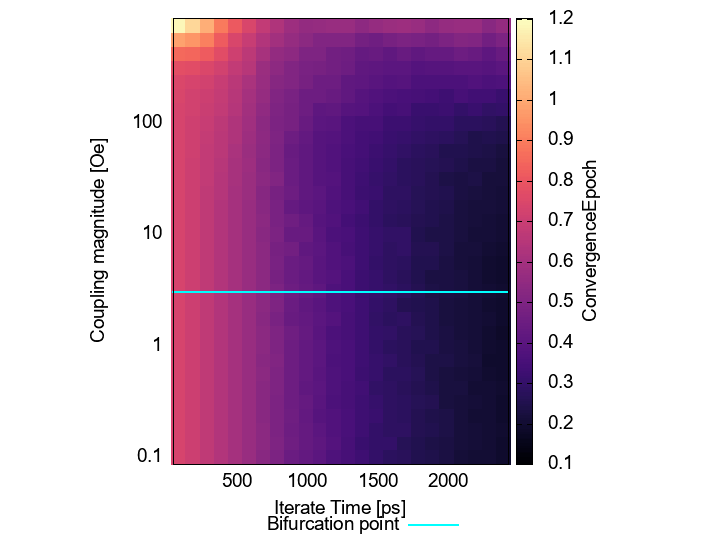}
    \end{center}
    \caption{
    The color map of average FTMLEs in the coupled STOs.
    The inputs of coupled STOs are MNIST data that are transformed by the trained input weight. 
    }
    \label{fig:sup_ftmle_csto}
\end{figure}

\begin{table*}[!t]
\caption{Coupled STO parameters}
\begin{center}
\scriptsize
\resizebox{0.8\textwidth}{!}{
  \begin{tabular}{lll} 
    Symbol & Explanation & Value \\  \midrule
    $M$ & saturation magnetization & 1448.3 emu/c.c. \\ 
    $H_{{\rm K}}$ & interfacial magnetic anisotropy field & 18.616 kOe \\ 
    $H_{{\rm appl}}$ & applied field & 200 Oe \\ 
    $V$ & volume of the free layer & $\pi \times 60^2 \times 2$ nm$^{3}$ \\ 
    $\eta$ & spin polarization & 0.537 \\ 
    $\lambda$ & spin-transfer torque asymmetry & 0.288 \\ 
    $\gamma$ &  gyromagnetic ratio & $1.764 \times 10^7$ rad/(Oe s) \\ 
    $\alpha$ & Gilbert damping constant & 0.005 \\ 
    $I$ & current & 2.5 mA \\ 
    ${\bf p}$ & unit vector in the direction of pinned layer magnetization & $(1, 0, 0)^{\top}$ 
  \end{tabular}
  }
  \label{tab:stoparam}
\end{center}
\end{table*}

We analyzed the FTMLEs of coupled STOs in detail and discuss the relationship between expansion property and information processing capabilities for our framework in this section.
We solved the MNIST task using coupled STOs under the same conditions as in Figure 5E and 5F of the main text and calculated the average value of FTMLEs for each input data. 
The results are shown in Figure \ref{fig:sup_ftmle_csto}.
The average FTMLEs were positive in all regions that were analyzed, and even in regions where the global dynamics of $A_cp < 2.0$ is non-chaotic, we observed that initial values that result in transient chaos were selected by learning.
The smaller the iteration time and the larger the coupling magnitude, the larger the average FTMLE.
The region with a large average FTMLE corresponded to the region with a small convergence epoch in Figure 5F of the main text. 
This suggests that the expansion property contributes to increased search efficiency in the stochastic gradient descent.

\subsection{Resistance to Noise in Coupled Spin-Torque Oscillators}
\label{sec:sup_csto_noise}

\begin{figure}
    \begin{center}
        \includegraphics[width=8cm]{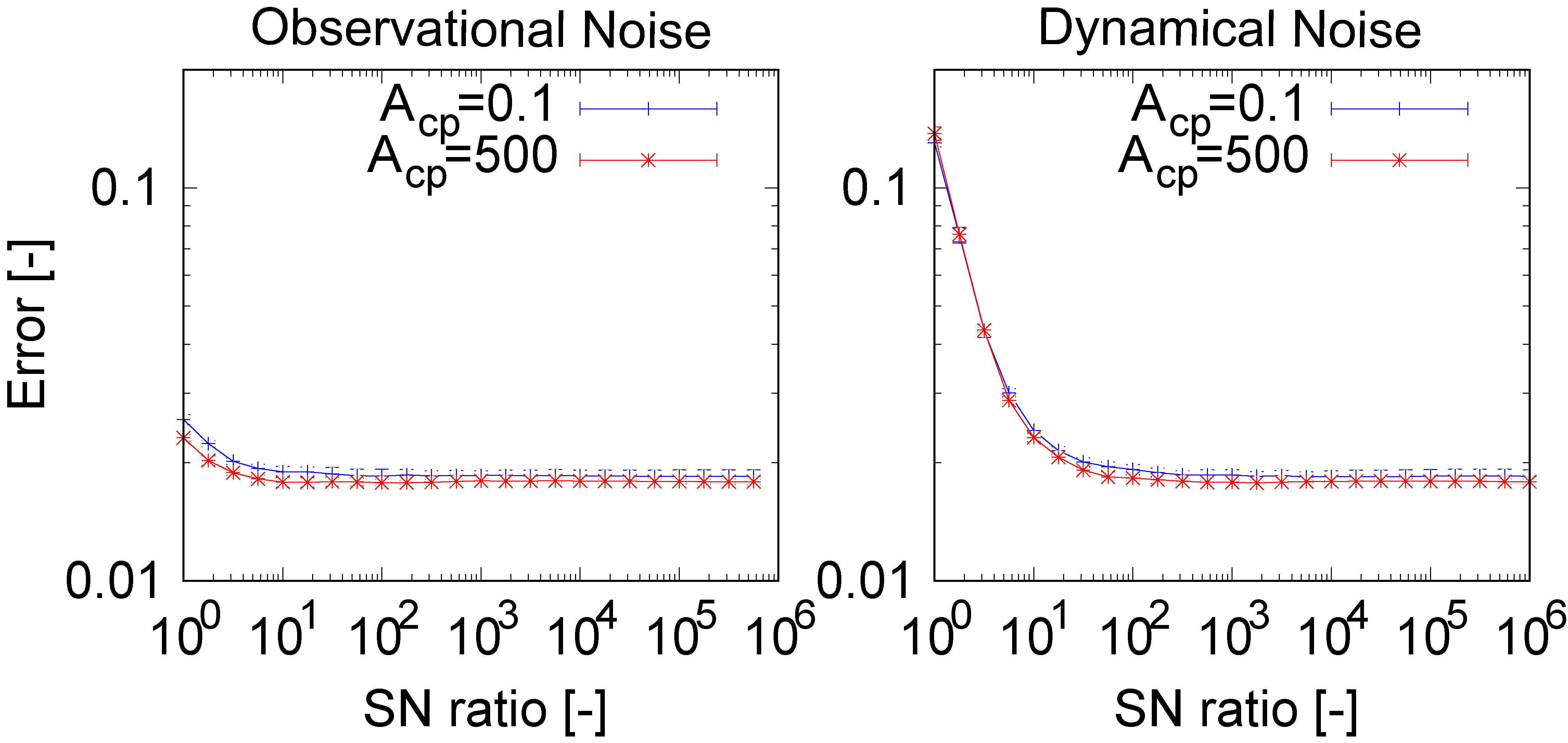}
    \end{center}
    \caption{
    The performances of the MNIST task using Coupled STOs with noises.
    The left- and right-hand sides show results with observational and dynamical noises, respectively.
    The vertical axes represent the errors $1 - {\text accuracy}$.
    }
    \label{fig:sup_csto_noise}
\end{figure}

In this section, we present the results of analyzing the robustness against noises in solving MNIST using Coupled STOs. Physical dynamics are inevitably subject to various types of noise, and noises in nonlinear systems can manifest as phenomena known as noise-induced phenomena, potentially exerting critical impacts on the system. Therefore, analyzing the robustness against noises is crucial when contemplating neuromorphic computations based on physical dynamics. Specifically, we analyzed the performance of the MNIST task using Coupled STOs when injecting two types of noise described by the following equations:
\begin{align}
x(0) &= f_{\rm norm}(W_{\rm in}u + {\bf p}_{\rm obs} )~, \\
x(i+1) &= f(W'x(i)) ~,~\text{and} \\
y &= W_{\rm out}(x(n)+ {\bf p}_{\rm dyn}) ~,
\end{align}

\noindent where ${\bf p}_{\rm obs}$ and ${\bf p}_{\rm obs}$ represent observation and dynamical noises, respectively, generated from Gaussian distributions. We conducted five trials by varying the signal-to-noise (SN) ratio $(A_s / A_n)$ under two conditions: coupling strength $A_{\rm cp} = 0.1$ and $500$ Oe, where $A_s$ and $A_n$ are variances of the state and noise, respectively.  The results are depicted in Figure \ref{fig:sup_csto_noise}. In both conditions, the performance change with respect to the SN ratio was continuous, confirming that noise did not have a fatal impact on information processing under observational and dynamic noises.

\subsection{Parameters of Coupled Spin-Torque Oscillators}
\label{sec:sup_csto_parameters}

In coupled STOs systems, the parameters specified within the Landau-Lifshitz-Gilbert equation, presented in Equations 17--18 in the main text, are detailed in Table~\ref{tab:stoparam}.


\end{document}